\documentclass{article}
\usepackage{booktabs}
\usepackage{caption}
\usepackage{placeins}
\usepackage{ntheorem}
\usepackage{siunitx}
\usepackage{tabularx}
\usepackage{changes}
\usepackage{tabu}
\usepackage{array,tabularx}
\usepackage{dblfloatfix}    % To enable figures at the bottom of page

\usepackage{threeparttablex}
\usepackage{subfigure}
\usepackage{graphicx} \usepackage{url}
\usepackage{hyperref}
\usepackage{url}
\usepackage{fancyhdr}
\usepackage{amsmath}
\usepackage{amssymb}
\usepackage[ruled,vlined]{algorithm2e}
\usepackage{multicol}
\usepackage{subfigure}
\usepackage{sidecap}
\usepackage{ulem}
\sidecaptionvpos{figure}{t}

\include{pythonlisting}
\usepackage[normalem]{ulem}
\usepackage[accepted]{icml2020}

\renewcommand*{\vec}[1]{\boldsymbol{#1}}
\newcommand{\DKL}{\mathbf{D}_\text{KL}}
\newcommand{\G}{\mathcal{G}}
\newcommand{\D}{\mathcal{D}}
\newcommand{\rpm}{\raisebox{.2ex}{$\scriptstyle\pm$}}

\icmltitlerunning{Implicit Generative Modeling for Efficient Exploration}

\begin{document}

\twocolumn[

\icmltitle{Implicit Generative Modeling for Efficient Exploration}

%\icmlsetsymbol{equal}{*}
\icmlsetsymbol{intern}{*}

\begin{icmlauthorlist}
    \icmlauthor{Neale Ratzlaff}{intern,osu}
    \icmlauthor{Qinxun Bai}{horizon}
    \icmlauthor{Li Fuxin}{osu}
    \icmlauthor{Wei Xu}{horizon}
\end{icmlauthorlist}

\icmlaffiliation{osu}{Department of Electrical Engineering and Computer Science, Oregon State University, Corvallis, Oregon, USA}
\icmlaffiliation{horizon}{Horizon Robotics, Cupertino, California, USA}

\icmlcorrespondingauthor{Neale Ratzlaff}{ratzlafn@oregonstate.edu}

\icmlkeywords{Machine Learning, Deep Reinforcement Learning, Reinforcement Learning, Bayesian Deep Learning, Uncertainty, Intrinsic Rewards}

\vskip 0.3in

\newcommand{\fix}{\marginpar{FIX}}
\newcommand{\new}{\marginpar{NEW}}
]
\printAffiliationsAndNotice{\icmlInternContribution}

\begin{abstract}
Efficient exploration remains a challenging problem in reinforcement learning, especially for those tasks where rewards from environments are sparse.
In this work, we introduce an exploration approach based on a novel implicit generative modeling algorithm to estimate a Bayesian uncertainty of the agent's belief of the environment dynamics. Each random draw from our generative model is a neural network that instantiates the dynamic function, hence multiple draws would approximate the posterior, and the variance in the predictions based on this posterior is used as an intrinsic reward for exploration.
We design a training algorithm for our generative model based on the amortized Stein Variational Gradient Descent.
In experiments, we demonstrate the effectiveness of this exploration algorithm in both pure exploration tasks and a downstream task, comparing with state-of-the-art intrinsic reward-based exploration approaches, including two recent approaches based on an ensemble of dynamic models. In challenging exploration tasks, our implicit generative model consistently outperforms competing approaches regarding data efficiency in exploration.

\end{abstract}

\section{Introduction}
\label{sec:intro}
Deep Reinforcement Learning (RL) has enjoyed recent success in a variety of applications, including super-human performance in Atari games~\citep{mnih2013atari}, robotic control~\citep{lillicrap2015continuous}, image-based control tasks~\citep{hafner2019planet}, and playing the game of Go~\citep{silver2016go}. 
Despite these achievements, many recent deep RL techniques still suffer from poor sample efficiency. Agents are often trained for millions, or even billions of simulation steps before achieving reasonable performance~\citep{burda2018large}. This lack of statistical efficiency makes it difficult to apply deep RL to real-world tasks, as the cost of acting in the real world is far greater than in a simulator. It is then a problem of utmost importance to design agents that make efficient use of collected data. In this work, we focus on efficient exploration which is widely considered to be one of the three key aspects in building a data-efficient agent
~\citep{sutton2018reinforcement}.

In particular, we focus on those challenging environments with sparse external rewards. In those environments, it is important for an effective agent to methodically explore a significant portion of the state space, since there may not be enough signals to indicate where the reward might be. Previous work usually utilize some sort of intrinsic reward
driven by the uncertainty in an agent's belief of the environment state~\citep{osband2018prior}. Intuitively, agents should explore more around states where they are not certain whether there could exist a previously unknown consequence -- which could be an unexpected extrinsic reward. However, uncertainty modeling from a deep network has proven to be difficult with no approach~\cite{snoek2019can} that is proven to be universally applicable.

In this work, we introduce a new framework of Bayesian uncertainty modeling for intrinsic reward-based exploration in deep RL. 
%Our framework characterizes the uncertainty in the agent's belief of the environment dynamics in a non-parametric manner to enable flexibility and expressiveness.
The main component of our framework is a network generator, each draw of which is a neural network that serves as the dynamic function for the environment. Multiple draws approximate a posterior of the dynamic model, and the variance in the future state predictions based on this posterior is used as an intrinsic reward for exploration. In doing so, our framework characterizes the uncertainty of the agent's belief of the environment dynamics in a non-parametric manner, avoiding restrictive distributional assumptions on the posterior, and explore a significantly larger model space than previous approaches. 
Recently, it has been shown~\citep{ratzlaff2019hypergan} that training these kinds of generators can be done in classification problems and the resulting network samples can represent a rich  distribution of diverse networks that perform approximately equally well on the classification task. 

For our goal of training this generator for the dynamic function, we propose 
a new algorithm to optimize the KL divergence between the implicit distribution (represented by draws from the generator) and the true posterior of the dynamic model (given the agent's experience) via amortized Stein Variational Gradient Descent (SVGD)~\citep{liu2016svgd,feng2017asvgd}. Amortized SVGD allows direct minimization of the KL divergence between the implicit posterior and true posterior without parametric assumptions or Evidence Lower Bound (ELBO) approximations, and projects to a finite-dimensional parameter update.

Comparing with recent work~\citep{pathak2019disagreement,shyam2019max} that maintain an ensemble of dynamic models and use the divergence or disagreement among them as an intrinsic reward for exploration, our implicit modeling of the posterior has two major advantages:
First, it is a more flexible framework for approximating the model posterior compared to an ensemble-based approximation. 
After one training episode, it can provide an unlimited amount of draws whereas for an ensemble each draw would require independent training.
Second, amortized SVGD~\citep{feng2017asvgd} allows direct nonparametric minimization of the KL divergence, in contrast with existing ensemble-based methods
that rely on the random initialization and/or bootstrapped experience sampling, which does not necessarily approximate the posterior.

In our experiments, we compare our approach with several state-of-the-art intrinsic reward-based exploration approaches, including two recent approaches that also leverage the uncertainty in dynamic models.
Experiments show that our implementation consistently outperforms competing methods regarding data efficiency in exploration.

In summary, our contributions are:
\begin{itemize}
\vspace{-0.1in}
    \item We propose a generative framework leveraging amortized SVGD to implicitly approximate the posterior of network parameters. Applying this framework to generate dynamic models of the environment, the uncertainty from the approximate posterior is used as an intrinsic reward for efficient exploration in deep RL.
    \vspace{-0.03in}
    \item We evaluate on three challenging exploration tasks and compare with three state-of-the-art intrinsic reward-based methods, two of which are also based on uncertainty in dynamic models. The superior performance of our method shows the effectiveness of the proposed framework in estimating the Bayesian uncertainty in the dynamic model for efficient exploration. We also evaluate in a dense reward setting to show its potential for improving downstream tasks.
\vspace{-0.1in}
\end{itemize}

\section{Problem Setup and Background}
\label{sec:background}

Consider a Markov Decision Process (MDP) represented as $(\mathcal{S}, \mathcal{A}, P, r, \rho_0)$, where $\mathcal{S}$ is the state space, $\mathcal{A}$ is the action space. $P:\mathcal{S}\times\mathcal{A}\times\mathcal{S}\to[0,1]$ is the unknown dynamics model, specifying the probability of transitioning to the next state $s'$ from the current state $s$ by taking the action $a$, as $P(s'|s,a)$. $r:\mathcal{S}\times\mathcal{A}\to\mathbb{R}$ is the reward function, $\rho_0:\mathcal{S}\to[0,1]$ is the distribution of initial states. A policy is a function $\pi: \mathcal{S}\times\mathcal{A}\to[0,1]$, 
which outputs a distribution over the action space for a given state $s$.

\subsection{Exploration in Reinforcement Learning}
\label{sec:explore}

In online decision-making problems, such as multi-arm bandits and reinforcement learning, a fundamental dilemma in an agent's choice is exploitation versus exploration. Exploitation refers to making the best decision given current information, while exploration refers to gathering more information about the environment. In the standard reinforcement learning setting where the agent receives an external reward
for each transition step, common recipes for exploration/exploitation trade-off include naive methods such as $\epsilon$-greedy~\citep{sutton2018reinforcement} and optimistic initialization~\citep{lai1985asymptotically}, posterior guided methods such as upper confidence bounds~\citep{auer2002ucb,dani2008ucb} and Thompson sampling~\citep{thompson1933likelihood}. 
We focus on the situation where external rewards are sparse or disregarded, here the above trade-off narrows down to the pure exploration problem of efficiently accumulating information about the environment. The common approach is to explore in a task-agnostic manner under some ``intrinsic" reward.
An exploration policy can then be trained with standard RL. 
Existing methods construct intrinsic rewards from visitation frequency of the state~\citep{bellemare2016unifying}, prediction error of the dynamic model as ``curiosity"~\citep{pathak2017curiosity}, diversity of visited states~\citep{eysenbach2018diversity}, etc.

\subsection{Dynamic Model Uncertainty as Intrinsic Reward}
\label{sec:intrinsic}

%Following the guiding principle of modeling Bayesian uncertainty in online decision making,
In order to model Bayesian uncertainty in online decision-making, two recent methods~\citep{pathak2019disagreement,shyam2019max} train an ensemble of dynamic models and use the variation/information gain as an intrinsic reward for exploration. In this work, we follow the similar idea of exploiting the uncertainty in the dynamic model, but emphasize the implicit posterior modeling in contrast with directly training an ensemble of dynamic models. 

Let $f: \mathcal{S}\times\mathcal{A} \rightarrow \mathcal{S}$ denote a model of the environment dynamics (represented by a neural network) we want to learn based on the agent experience $\mathcal{D}$. We design a generator module $\G$ which takes a random draw from the standard normal distribution and outputs a sample vector of parameters $\vec{\theta}$ that determines $f$ (denoted as $f_{\vec{\theta}}$). 
If samples from $\G$ represent the posterior distribution $p(f_{\vec{\theta}}|D)$, then given $(s_t, a_t)$, the uncertainty in the output of the dynamics model can be computed by the following variance among a set of samples $\{\vec{\theta}_i\}_{i=1}^m$ from $\G$, and used as an intrinsic reward $r^{in}$ for learning an exploration policy,
\begin{equation}
    \label{eq:var_reward}
    r^{in}_t=\frac{1}{m}\sum\nolimits_{i=1}^m\left\|f_{\vec{\theta}_i}(s_t,a_t) - \frac{1}{m}\sum\nolimits_{\ell=1}^m f_{\vec{\theta}_\ell}(s_t,a_t)\right\|^2.
\end{equation}
When training the exploration policy, this intrinsic reward can be computed with rollouts in the environment, or simulated rollouts generated by the estimated dynamic model.

\section{Posterior Approximation via Amortized SVGD}
\label{sec:svgd}

In this section, we introduce the core component of our exploration agent, the dynamic model generator $\G$. In the following subsections, we first introduce the design of this generator and then describe its training algorithm in detail. A summary of our algorithm is given in the last subsection.

\subsection{Implicit Posterior Generator}
\label{sec:generator}

\begin{figure}[t]
    \centering
    \includegraphics[width=.75\linewidth]{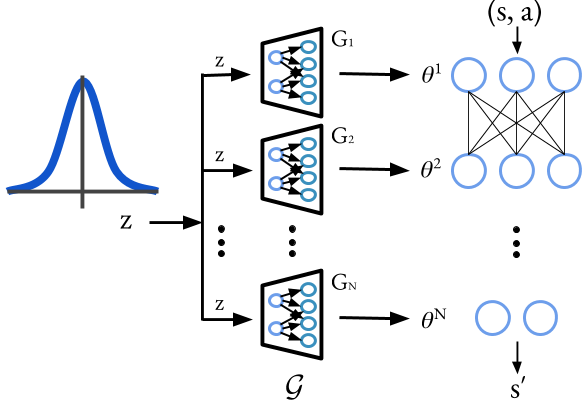}
    \vskip -0.07in
    \caption{Architecture of the layer-wise generator of the dynamic model. A single shared noise sample $z \in \mathbb{R}^d$ is drawn from a standard Gaussian with diagonal covariance, and input to layer-wise generators $\{G_1,\cdots,G_N\}$. Each generator $G_j$ outputs parameters $\theta^j$ for the corresponding $j$-th layer of the neural network representing the dynamic model.} \label{fig:hypergan}
    \vskip -0.1in
\end{figure}
As shown in Fig.~\ref{fig:hypergan}, the dynamic model is defined as an $N$-layer neural network function $f_{\vec{\theta}}(s,a)$, with input (state, action) pair $(s,a)$ and model parameters $\vec{\theta}=(\theta^1,\cdots,\theta^N)$, where $\theta^j$ represents network parameters of the $j$-th layer. The generator module $\G$ consists of exactly $N$ layer-wise generators, $\{G_1, \cdots, G_N\}$, where each $G_j$ takes the random noise vector $z\in\mathbb{R}^d$ as input, and outputs the corresponding parameter vector $\theta^j = G_j(z;\eta^j)$, where $\eta^j$ are the parameters of $G_j$. Note that $z$ is sampled from a $d$-dimensional standard normal distribution, and is shared across all generators to capture correlations between the generated parameters.  
%independently from a $d$-dimensional standard normal distribution.%, rather than jointly. 
As mentioned in Sec.~\ref{sec:intro}, this framework has advantages in flexibility and efficiency, comparing with ensemble-based methods~\citep{shyam2019max,pathak2019disagreement},
since it maintains only parameters of the $N$ generators, i.e., $\vec{\eta}=(\eta^1, \cdots,\eta^N)$, and enables drawing an arbitrary number of sample networks to approximate the posterior of the dynamic model.

\subsection{Training with Amortized SVGD}
\label{sec:svgd}

We now introduce the training algorithm of the generator module $\G$. 
Assuming that the true posterior of the dynamic model given the agent's experience $\D$ is $p(f|\D)$, 
%and the implicit distribution of $f_{\vec{\theta}}$ captured by $\G$ is $q(f_{\vec{\theta}})$. 
and the implicit distribution captured by $\G$ is $q(f_{\vec{\theta}})$, where $f_{\vec{\theta}}$ denotes the function values obtained by evaluating $f_{\vec{\theta}}$ on $\D$. 
We want $q(f_{\vec{\theta}}|\D)$ to be as close as possible to $p(f|\D)$, such closeness is 
commonly measured by the KL divergence $\DKL\left[q(f_{\vec{\theta}}|\D)\|p(f|\D)\right]$. 
The traditional approach for finding $q$ that minimizes $\DKL\left[q(f_{\vec{\theta}}|\D)\|p(f|\D)\right]$ is variational inference (VI), by maximizing the ELBO \cite{blei2017variational}. But standard VI necessitates restricting the parametric form of the target posterior. Recently, a nonparametric VI framework, Stein Variational Gradient Descent (SVGD)~\citep{liu2016svgd}, was proposed, 
that represents $q$ with a set of particles rather than making any parametric assumptions, and approximates the functional gradient descent w.r.t. $\DKL\left[q(f_{\vec{\theta}}|\D)\|p(f|\mathcal{D})\right]$ by iterative particle evolvement.
We apply SVGD to our sampled network functions, and follow the idea of amortized SVGD~\citep{feng2017asvgd} to project the functional gradients to the parameter space of $\vec{\eta}$ by back-propagation through the generators.

Given a set of dynamic functions $\{f_{\vec{\theta}_i}\}_{i=1}^m$ sampled from $\G$, SVGD updates each function by 
\begin{equation*}
    f_{\vec{\theta}_i} \leftarrow f_{\vec{\theta}_i} + \epsilon \phi^*(f_{\vec{\theta}_i}), \qquad i=1, \cdots, m,
\end{equation*}
where $\epsilon$ is step size, and $\phi^*$ is the function in the unit ball of a reproducing kernel Hilbert space (RKHS) $\mathcal{H}$ that maximally decreases the KL divergence between the distribution $q$ represented by $\{f_{\vec{\theta}_i}\}_{i=1}^m$ and the target posterior $p$.
% Fuxin: This abuse is OK
%----------\textcolor{red}{Neale}\\
% Instead of considering the functions directly, we abuse notation slightly and denote by $f_{\vec{\theta}}$ the values given by evaluating $f_{\vec{\theta}}$ at data points $x$. 

% Original (in paper) equation
%\begin{equation*}
%    \phi^* = \underset{\phi \in \mathcal{H}}{\max} \left\{ -\frac{d}{d\epsilon} \DKL(q || p), \quad s.t. ||\phi||_\mathcal{H} \leq 1
%    \right\}.
%\end{equation*}

Let $q_{[\epsilon \phi]}$ refer to the distribution of updated particles. The optimal choice of $\phi$ can be found by solving the following optimization problem:

\begin{equation}
    \phi^* = \underset{\phi \in \mathcal{H}}{\max} \left\{ -\frac{d}{d\epsilon} \DKL(q_{[\epsilon \phi]} || p), \quad s.t. ||\phi||_\mathcal{H} \leq 1
    \right\}.
 \label{eq:kl_optim}
\end{equation}

It was shown in \cite{liu2016svgd} that the KL-divergence can be expressed as a linear functional of $\phi$, 

\begin{equation}
    -\frac{d}{d\epsilon} \DKL(q_{[\epsilon \phi]} || p)|_{\epsilon=0} = \mathbb{E}_{f_{\vec{\theta}} \sim q} \left[ \mathcal{S}_p \phi(f_{\vec{\theta}}) \right]
    \label{eq:stein}
\end{equation}

where $\mathcal{S}_p$ is the Stein operator \citep{stein2004use}:
\begin{equation*}
   \mathcal{S}_p \phi( f_{\vec{\theta}} )= \nabla_{f_{\vec{\theta}}} \log p(f_{\vec{\theta}})^T \phi(f_{\vec{\theta}}) + \nabla_{f_{\vec{\theta}}}^T \phi(f_{\vec{\theta}})
\end{equation*}
%According to eq. (\ref{eq:kl_optim}), the Stein operator must return zero if the distributions $p$ and $q$ match. 

%\begin{equation*}
%    \mathbb{E}_p\left[\mathcal{S}_p \phi \right] = \mathbb{E}_p\left [\nabla_{f_{\vec{\theta}}} \log p^T \phi + \nabla_{f_{\vec{\theta}}} \cdot \phi \right] = 0
%\end{equation*}
%We can equivalently express \ref{eq:kl_optim} as a search over functions in $\mathcal{H}$, to find a $\phi^*$ that is maximal under the Stein operator. 
%\begin{equation}
%    \phi^* = \underset{\phi \in \mathcal{H}}{\max} \left\{ \mathbb{E}_{f_{\vec{\theta}} \sim q} \mathcal{S} \phi(f_{\vec{\theta}}), \quad s.t. ||\phi||_\mathcal{H} \leq 1
%    \right\}.
% \label{eq:stein_optim}
%\end{equation}

%-------\textcolor{red}{End Neale}\\

%Utilizing the kernel trick to express $\phi(f_{\theta_i})$ in $span\{k(\cdot, f_{\theta_i}), i = 1, \ldots, N\}$, 
Hence, eq. (\ref{eq:kl_optim}) has a closed form solution,
\begin{equation}
    \phi^*(f_{\vec{\theta}}) = \underset{f_{\theta} \sim q}{\mathbb{E}} \left[ \nabla_{f_{\vec{\theta}}} \log p(f_{\vec{\theta}}) k(f_{\vec{\theta}}, f_{\vec{\theta}_i}) + \nabla_{f_{\vec{\theta}}} k(f_{\vec{\theta}}, f_{\vec{\theta}_i}) \right],
    \label{eq:phi}
\end{equation}

where $k(\cdot, \cdot)$ is the positive definite kernel associated with the RKHS.
The log-likelihood term for $f_{\vec{\theta}}$ corresponds to the negation of the regression loss of future state prediction for all transitions in $\D$, i.e., 
$\log p(f_{\vec{\theta}})
=-\sum_{(s,a,s')\in \D}L(f_{\vec{\theta}}(s,a), s')$.
Given that each $\vec{\theta}_i$ is generated by $\G(\vec{z};\vec{\eta})$, 
the update rule for $\vec{\eta}$ can be obtained by the chain rule, 
%\begin{equation}
%\label{eq:eta}
%    \vec{\eta} \leftarrow \vec{\eta} + \epsilon\sum_{i=1}^m\nabla_{\vec{\eta}}\G(\vec{z}_i; \vec{\eta})\phi^*(\vec{\theta}_i),
%\end{equation}
%(\textcolor{red}{Neale: Change eq 6 to})
\begin{equation}
    \label{eq:eta}
    \vec{\eta} \leftarrow \vec{\eta} + \epsilon\sum_{i=1}^m \nabla_{\vec{\eta}} \phi^*(f_{\vec{\theta}_i})|_{\vec{\theta}_i=\G(\vec{z}_i; \vec{\eta})}
\end{equation}
where $\phi^*(\G(\vec{z}_i; \vec{\eta}))$ can be computed by (\ref{eq:phi}) using empirical expectation from sampled batch $\{\vec{\theta}_i\}_{i=1}^m$,
\begin{equation}
\begin{split}
    \phi^*(f_{\vec{\theta}_i}) = \frac{1}{m}\sum_{\ell=1}^m\left\{-\left[\sum\nolimits_{(s,a,s')\in \D}
    \nabla_{f_{\vec{\theta}_\ell}}L(f_{\vec{\theta}_\ell}(s,a), s')\right] \right. \\
    \cdot \left. k(f_{\vec{\theta}_\ell(s,a)}, f_{\vec{\theta}_i(s,a)}) 
    + \nabla_{f_{\vec{\theta}_\ell}} k(f_{\vec{\theta_\ell}(s,a)}, \vphantom{\sum_{\ell=1}^m} f_{\vec{\theta}_i(s,a)})\right \},
    \label{eq:phi_emp}
\end{split}
\end{equation}
where $k(\cdot, \cdot)$ is the Gaussian kernel evaluated at function outputs, which is in the state space.

\subsection{Summary of the Exploration Algorithm}
\label{sec:sum_alg}
To condense what we have proposed so far, we summarize in Algorithm~\ref{algo:implicit-active-ex} the procedure used to train the generator of dynamic models and the exploration policies.

\begin{algorithm}[H]
    \SetAlgoLined
    \textbf{Initialize} Generator $\G_{\vec{\eta}}$, parameters $T,m$ \\\textbf{Initialize} Policy $\pi$, Experience buffer $\D$
    \\
    \While{\normalfont{True}}{
\While{\normalfont{episode not \textbf{done}}:} { 
$f_{\vec{\Theta}} \leftarrow \mathcal{G}(\vec{z};\vec{\eta}), \vec{z}\sim\mathcal{N}(0, I^d)$ \\
$\vec{\eta} \leftarrow$ \normalfont{evaluate (\ref{eq:eta}),~(\ref{eq:phi_emp}) on $\D$} \\
$\widetilde{\D}\sim\text{MDP}(f_{\vec{\Theta}}) \\ \D_{\pi} \leftarrow \D \cup\widetilde{\D}$,\\
            $R_{\pi} \leftarrow r^{in}(f_{\vec{\theta}},s,a|(s,a)\sim \D_{\pi})\ \text{by}\ (\ref{eq:var_reward})$ \\
            $\pi \leftarrow \text{update policy on}\ (\D_{\pi},R_{\pi})$ \\
            $\D_T \leftarrow \text{rollout}\ \pi\ \text{for} \ T\ \text{steps}$ \\
$\D \,\, \leftarrow \D \cup \D_T$
    
        }
     }
    \caption{Exploration with an Implicit Distribution}
    \label{algo:implicit-active-ex}
\end{algorithm}
\vspace{-0.05in}
Our algorithm starts with a buffer $\D$ of random transitions and explores for some fixed number of episodes. For each episode, our algorithm samples a set of dynamic models $f_{\vec{\Theta}}=\{f_{\vec{\theta}_i}\}$ from the generator $\G$, and updates the generator parameters $\vec{\eta}$ using amortized SVGD (\ref{eq:eta}) and (\ref{eq:phi_emp}). 
For the policy update, the intrinsic reward (\ref{eq:var_reward}) is evaluated on the actual experience $\D$ and the simulated experience $\widetilde{\D}$ generated by $f_{\vec{\theta}_i}$.
The exploration policy is then updated using a model-free RL algorithm on the collected experience $\D_{\pi}$ and intrinsic rewards $R_{\pi}$.
The updated exploration policy is then used to rollout in the environment for $T$ steps so that new transitions are collected and added to the buffer $\D$. The process is repeated until the end of the episode.% repeat the process, until the episode is done.

\section{Related Work}
Efficient exploration remains a major challenge in deep reinforcement learning~\citep{fortunato2017noisy,burda2018exploration,eysenbach2018diversity,burda2018large}, and there is no consensus on the \textit{correct} way to explore an environment. One practical guiding principle for efficient exploration is the reduction of the agent's epistemic uncertainty of the environment~\citep{chaloner1995bayesian, osband2017value}.
~\citet{osband2016bootdqn} uses a bootstrap ensemble of DQNs, where the predictions of the ensemble are used as an estimate of the agent's uncertainty over the value function. 
~\citet{osband2018prior} proposed to augment the predictions of a DQN agent by adding the contribution from a prior 
to the value estimate. In contrast to our method, these approaches seek to estimate the uncertainty in the \textit{value function}, while we focus on exploration with intrinsic reward by estimating the uncertainty of the \textit{dynamic model}.
~\citet{fortunato2017noisy} add parameterized noise to the agent's weights, to induce state-dependant exploration beyond $\epsilon$-greedy or entropy bonus.  

Methods for constructing intrinsic rewards for exploration have become the subject of increased study. 
One well-known approach is to use the prediction error of an inverse dynamics model as an intrinsic reward ~\citep{pathak2017curiosity, schmidhuber1991curious}.
~\citet{schmidhuber1991curious} and ~\citet{sun2011planning} proposed using the learning progress of the agent as an intrinsic reward.
Count based methods~\citep{bellemare2016unifying,ostrovski2017count} give a reward proportional to the visitation count of a state. 
~\citet{HouVIME} formulate exploration as a variational inference problem, and use Bayesian neural networks (BNN) to maintain the agent's belief over the transition dynamics. The BNN predictions are used to estimate a form of Bayesian information gain called compression improvement.   The variational approach is also explored in ~\citet{mohamed2015variational,gregor2016variational,salge2014empowerment}, who proposed using intrinsic rewards based on a variational lower bound on empowerment: the mutual information between an action and the induced next state. This reward is used to learn a set of discriminative low-level skills. 
The most closely-related work to ours are two recent methods~\citep{pathak2019disagreement,shyam2019max} that compute intrinsic rewards from an ensemble of dynamic models. Disagreement among the ensemble members in next-state predictions is computed as an intrinsic reward. 
\citet{shyam2019max} also uses \textit{active exploration}~\citep{schmidhuber2003exploring, chua2018deep}, in which the agent is trained in a surrogate MDP, to maximize intrinsic reward before acting in the real environment. 
Our method follows the similar idea of exploiting the uncertainty in the dynamic model, but instead suggests an implicit generative modeling of the posterior of the dynamic function, 
which enables a more flexible approximation of the posterior uncertainty with better sample efficiency.

There has been a wealth of research on nonparametric particle-based variational inference methods~\citep{liu2016svgd,dai2016provable,ambrogioni2018wasserstein}, where particles are maintained to represent the variational distribution, and updated by solving an optimization problem within an RKHS.
Notably, we use amortized SVGD~\citep{feng2017asvgd} to optimize our generator for approximately sampling from the posterior of the dynamic model. In addition to amortized SVGD, other nonparametric methods for training implicit samplers with particle-based variational inference have been proposed, such as ~\citet{li2018gradient}.  

\section{Experiments}
\label{sec:exp_experiments}
In this section we conduct experiments to compare our approach to the existing state-of-the-art in efficient exploration with intrinsic rewards to illustrate the following:
\begin{itemize}
\vspace{-0.05in}
    \item An agent with an implicit posterior over dynamic models explores more effectively and efficiently than agents using a single model or a static ensemble. 
    \item Agents seeking external reward find better policies when initialized from powerful exploration policies. 
    Our ablation studies shows that the better the exploration policy as an initialization, the better the downstream task policy can learn.
\vspace{-0.05in}
\end{itemize} 

To evaluate the proposed method in terms of exploration efficiency,
we first consider exploration tasks agnostic of any external reward. 
In this setting, the agent explores the environment irrespective of any downstream task. 
Then, to further investigate the potential of our exploration policies, we consider 
transferring the learned exploration policy to downstream task policies where a dense external reward is provided.  Note that both cases are important for understanding and applying exploration policies.
In sparse reward settings, such as a maze, the reward could occur at any location, without informative hints accessible at other locations.
Therefore an effective agent must be able to efficiently explore the entire state space in order to consistently find rewards under different task settings.
In dense reward settings, the trade-off between exploration and exploitation plays a central role in efficient policy learning. 
Our experiments show that 
even for a state-of-the-art model-free algorithm like Soft Actor-Critic (SAC) \cite{haarnoja2018soft}, that already incorporates a strong exploration mechanism, %(the maximum entropy framework),
spending some initial rollouts to learn a powerful exploration policy as an initialization of the task policy still considerably improves the learning efficiency.

\subsection{Pure Exploration Results}
\label{sec:pure_explore}
For pure exploration,
we consider three challenging continuous control tasks in which efficient exploration is known to be difficult. In each environment, the dynamics are nonlinear and cannot be solved with tabular approaches. As explained in the beginning of Section~\ref{sec:exp_experiments}, the agent does not receive any external reward and is motivated purely by the uncertainty in its belief of the environment.

\noindent {\bf Experimental setup}
To validate the effectiveness of our method, we compare with several state-of-the-art formulations of intrinsic reward. Specifically, we conduct experiments comparing the following methods: 
\begin{itemize}
    \vspace{-0.07in}
  \setlength{\itemsep}{-4pt}
    \item (\emph{Ours}) The proposed intrinsic reward, using the estimated variance from an implicit distribution of the dynamic model. 
    \item (\emph{Random}) Random exploration as a naive baseline.
    \item (\emph{ICM}) Error between predicted next state and observed next state~\citep{pathak2017curiosity}.
    \item (\emph{Disagreement}) Variance of predictions from an ensemble of dynamic models~\citep{pathak2019disagreement}.
    \item (\emph{MAX}) Jensen-Renyi information gain of the dynamic function~\citep{shyam2019max}.
    \vspace{-0.07in}
\end{itemize}

\noindent {\bf Implementation details}

Since our goal is to compare the performance across different intrinsic rewards, we fix the model architecture, training pipeline, and hyper-parameters across all methods,\footnote{We use the codebase of \emph{MAX} as a basis and implement \emph{Ours}, \emph{ICM}, and \emph{Disagreement} intrinsic rewards under the same framework. The full \emph{Disagreement} method includes an additional differentiable reward function that we compare with separately in the supplementary material.}
shared hyper-parameters follow the \emph{MAX} default settings.
For the purpose of computing the information gain, dynamic models for \emph{MAX} predict both mean and variance of the next state, while for other methods, dynamic models predict only the mean. 
Since our method trains a generator of dynamic models instead of a fixed-size ensemble, we fix the number of models we sample from the generator at $m=32$, 
which equals the ensemble size for \emph{MAX}, and \emph{Disagreement}. 
For all experiments except for the Chain environment, we use SAC v1 \cite{haarnoja2018soft} as the model-free RL algorithm used to train the exploration policies. 
\begin{figure}[ht!]
    \centering
    \includegraphics[width=1.0\linewidth]{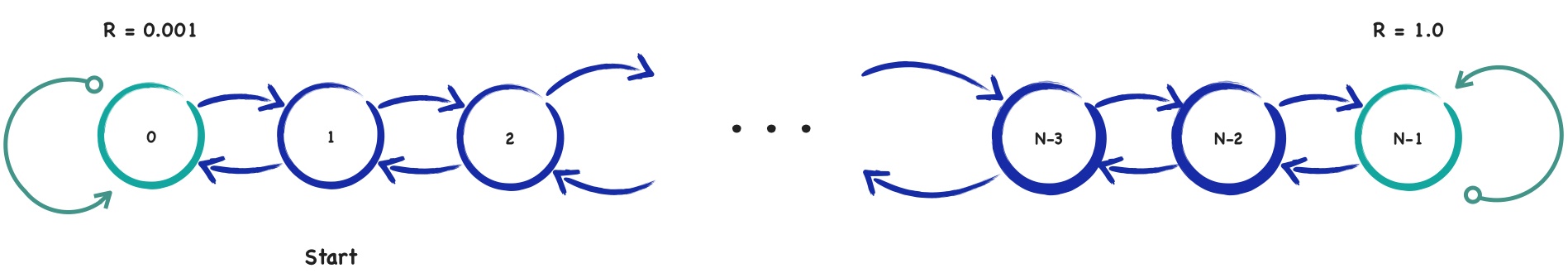}
    \vspace{-1em}
    \caption{The NChain environment.}
    \label{fig:chain}
    \vskip -0.05in
\end{figure}

\subsubsection{Toy Task: NChain} 

\begin{figure}[ht!]
    \centering \vspace{-2mm}
    \includegraphics[width=1.0\linewidth]{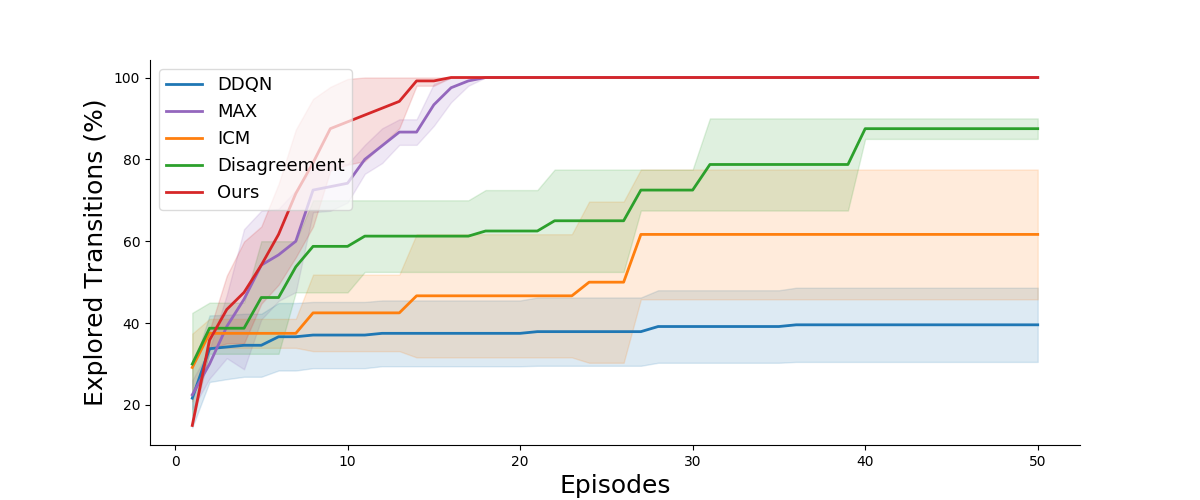}
    \vskip -0.05in
    \caption{Results on the 40-link chain environment. Each line is the mean of three runs, with the shaded regions corresponding to $\pm 1$ standard deviation. Our method and \emph{MAX} actively reduce uncertainty in the chain, and are able to quickly explore to the end of the chain. $\epsilon$-greedy \emph{DDQN} fails to explore more than 40\% of the chain. Both \emph{ICM} and \emph{Disagreement} perform better than \emph{DDQN} but explore less efficiently compared to \emph{MAX} and our method}
    \label{fig:chain_res}
    \vskip -0.05in
\end{figure}

As a sanity check, 
we first follow \emph{MAX}~\citep{shyam2019max} by evaluating our method on a stochastic version of the toy environment NChain. As shown in Fig.~\ref{fig:chain}, the chain is a finite sequence of $N$ states. 
Each episode starts from state $1$ and lasts for $N+9$ steps. For each step, the agent can move forward to the next state in the chain or backward to the previous state. 
Attempting to move off the edge of the chain results in the agent staying still. 
Reward is only afforded to the agent at the edge states: $0.01$ for reaching state $0$, and $1.0$ for reaching state $N-1$. In addition, there is uncertainty built into the environment: each state is designated as a \textit{flip-state} with probability $0.5$. When acting from a flip-state, the agent's actions are reversed, i.e., moving forward will result in movement backward, and vice-versa. Given the (initially) random dynamics and a sufficiently long chain, we expect an agent 
using an $\epsilon$-greedy exploration strategy to exploit only the small reward of state $0$. In contrast, agents with exploration policies which actively reduce uncertainty can efficiently discover all states in the chain. Fig.~\ref{fig:chain_res} shows that our agent navigates the chain in less than 15 episodes, while the $\epsilon$-greedy agent (double DQN) does not make meaningful progress. We also evaluate each of the methods enumerated in section \ref{sec:pure_explore}. 

We find that actively reducing uncertainty is critical to exploring the chain. We believe that because \emph{ICM} explores using the prediction error of the dynamic model, a chain initialized with simple dynamics (few flip states) may lead to poor exploration. Though \emph{Disagreement} uses a similar intrinsic reward as \emph{Ours}, we suspect the use of a static ensemble leads to a lack of predictive diversity, as the ensemble can easily overfit to the dynamics of the chain, limiting exploration. \emph{MAX} may avoid overfitting to the chain due to using stochastic neural networks. Our method however, directly promotes model diversity using amortized SVGD, and uses the uncertainty in our dynamic model to explore new states. We provide additional details of the NChain experiments in the supplementary material.  

\setcounter{figure}{5}
\begin{figure*}[hb]
\centering 
    \subfigure[Ant Maze]{
        \includegraphics[width=.18\linewidth]{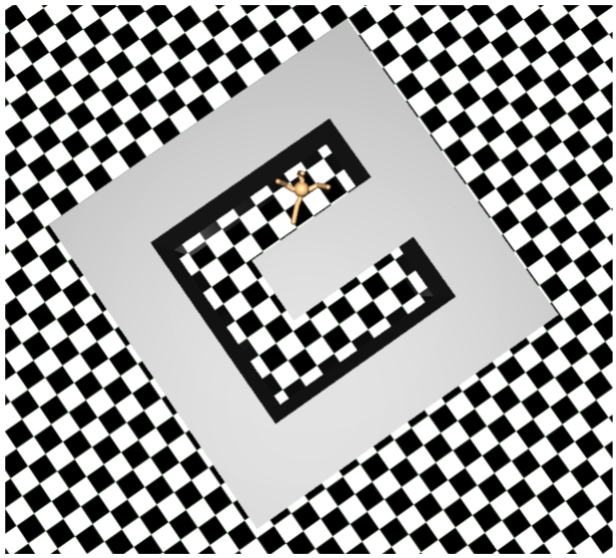}
        \label{fig:ant_maze}
    }
    \subfigure[2500 Steps]{
        \vspace{-1cm}
        \includegraphics[width=.17\linewidth]{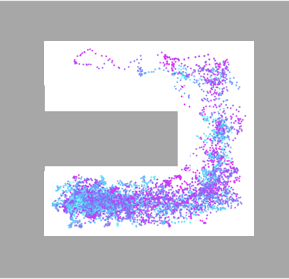}
        \label{fig:state_visit1}
    }
    \subfigure[5000 Steps]{
        \vspace{-1cm}
        \includegraphics[width=.17\linewidth]{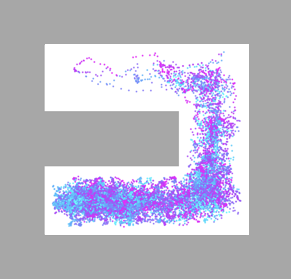}
        \label{fig:state_visit2}
    }
    \subfigure[7500 Steps]{
        \vspace{-1cm}
        \includegraphics[width=.17\linewidth]{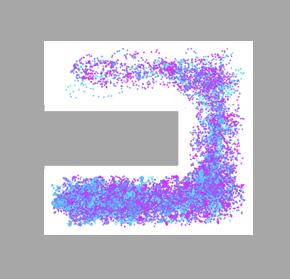}
        \label{fig:state_visit3}
    }
    \subfigure[10000 Steps]{
    
        \vspace{-1cm}
        \includegraphics[width=.20\linewidth]{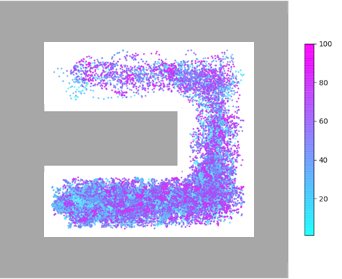}
        \label{fig:state_visit4}
    }
\caption{Figure (a) displays U-shaped ant maze. Figures (b-e) show the behavior of the agent at different stages of training, over 5 seeds. Points are color-coded with blue points occurring at the beginning of the episode, and red points at the end.
}
\label{fig:state_visit}
\end{figure*}
\setcounter{figure}{3}

\subsubsection{Acrobot Control}

\begin{figure}[ht!]
    \centering 
    \includegraphics[width=1.0\linewidth]{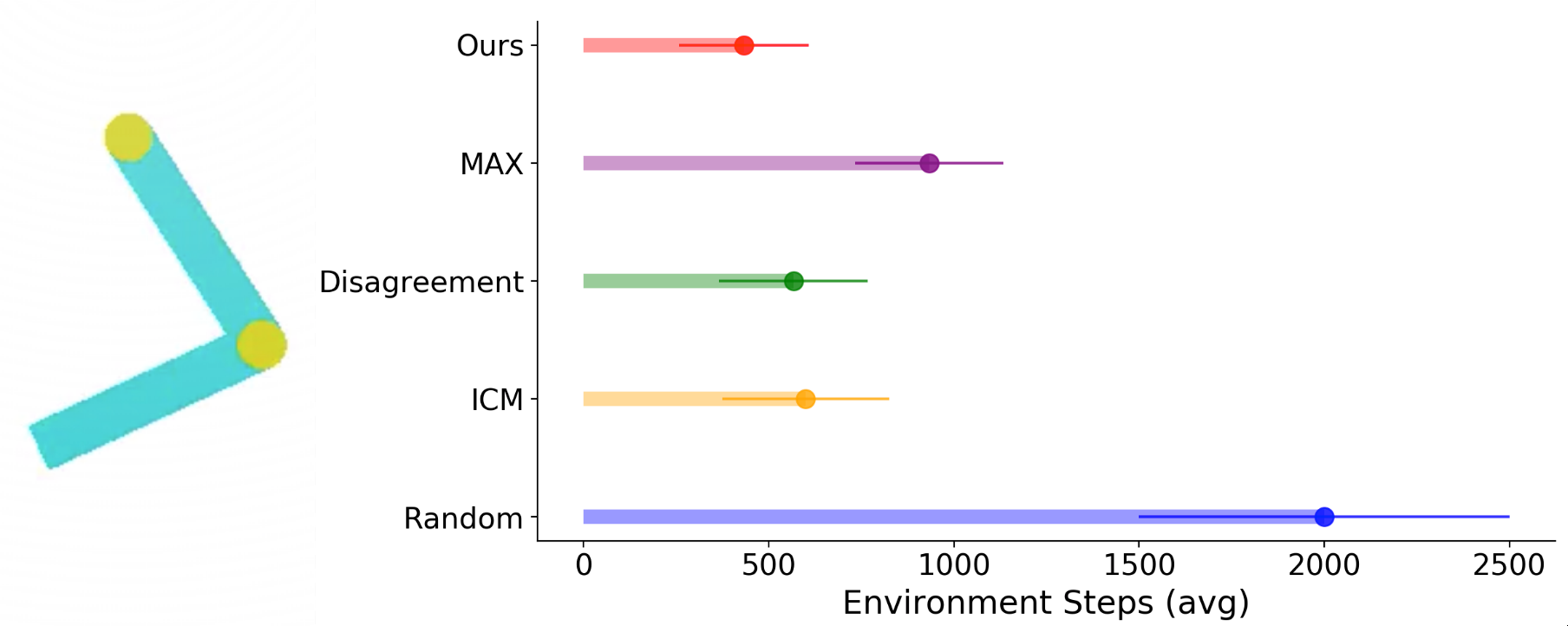}
    \vskip -0.05in
    \caption{Performance of each method on the Acrobot environment (average of five seeds), with error bars representing $\rpm 1$ standard deviation. The length of each horizontal bar indicates the number of environment steps each agent/method takes to swing the acrobot to fully horizontal on both (left and right) directions.}
    \label{fig:control_results}
\vskip -0.05in
\end{figure}

The first continuous control environment that we evaluate is a modified version of the Acrobot. As shown in figure~\ref{fig:control_results}, the Acrobot environment begins with a hanging down pendulum which  
consists of two links connected by an actuated joint. Normally, 
a discrete action $a \in \{-1, 0, 1\}$ either applies a unit force on the joint in the left or right direction $(a=\pm 1)$, or not $(a = 0)$. We modify the environment such that a continuous action $a \in [-1, 1]$ applies a force $F = |a|$ in the corresponding direction.

To focus on efficient exploration, we test the ability of each exploration method to sweep the entire lower hemisphere: positioning the acrobot completely horizontal towards both (left and right) directions. 
Given this is a relatively simple task and can be solved by random exploration, as shown in Figure~\ref{fig:control_results}, all four intrinsic reward methods solve it within just hundreds of steps
and our method is the most efficient one.
The takeaway here is that in relatively simple environments where there might be little room for improvement over state-of-the-art, our method still achieves a better performance due to its flexibility and efficiency in approximating the model posterior. As we will see in subsequent experiments, this observation scales well with the increasing difficulty of the environments. 

\begin{figure}[ht!]
    \subfigure[Ant Navigation Task Results]{
        \label{fig:nav_results}
        \includegraphics[width=1.0\linewidth]{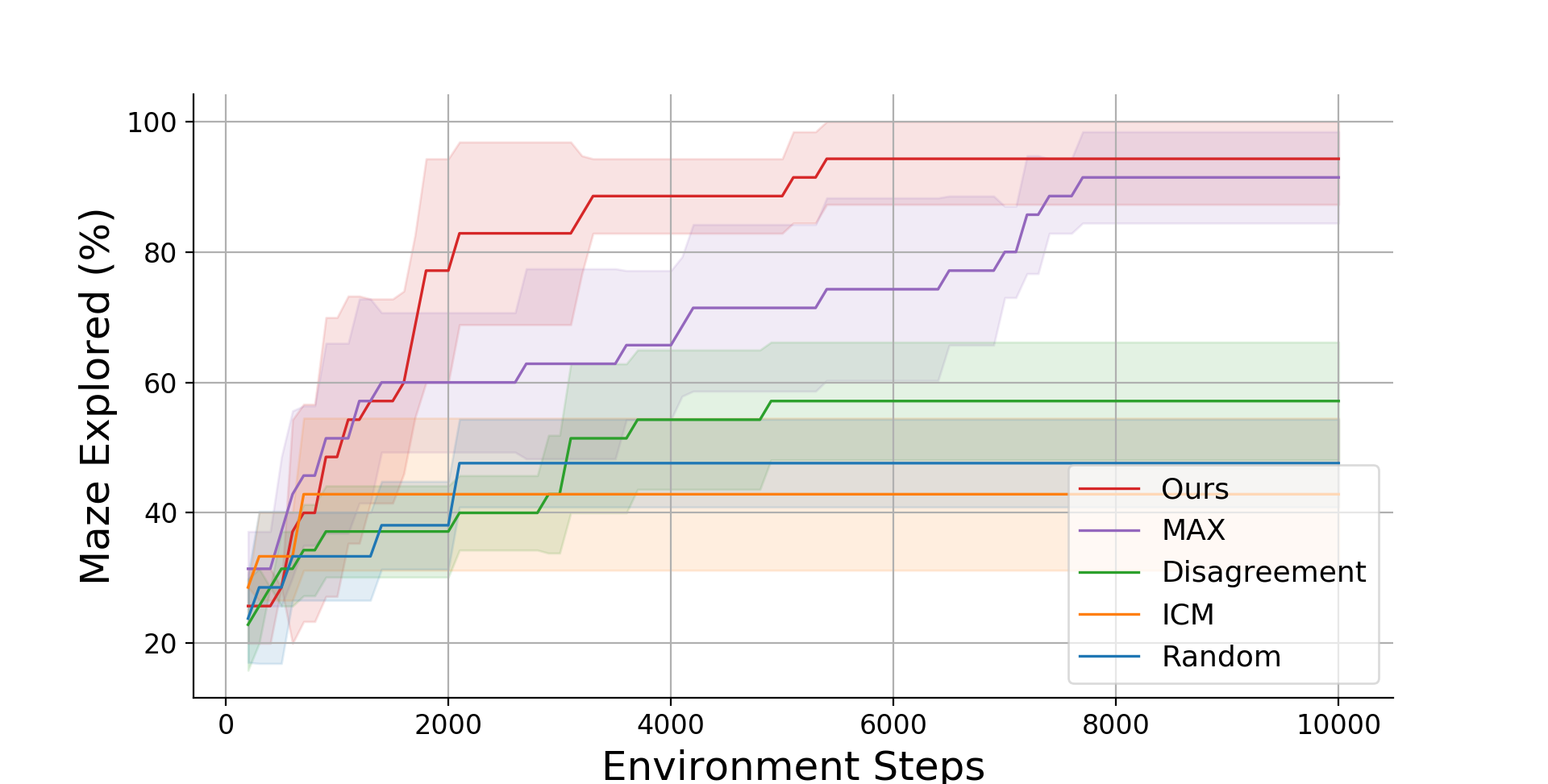}
    }
    \subfigure[Ant Intrinsic Rewards]{
    \vspace{-0.1in}
        \label{fig:ant_int_reward}
        \hspace{0.025in}
        \includegraphics[width=0.9\linewidth]{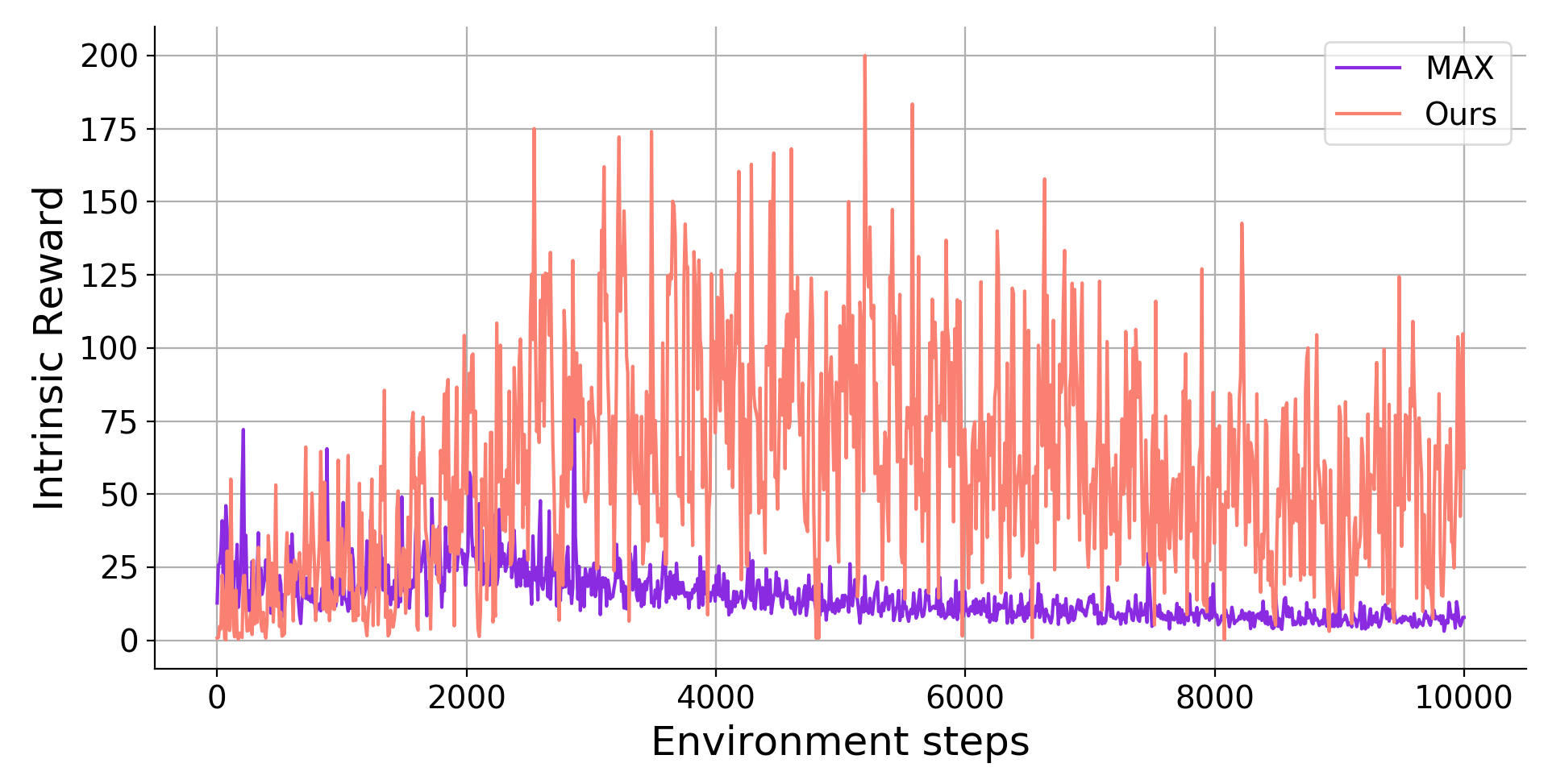}
    }
    \vskip -0.1in
\caption{Figure (a) shows the performance of each method with mean and $\pm 1$ standard deviation (shaded region) over five seeds. $x$-axis is the number of steps the ant has moved, $y$-axis is the percentage of the U-shaped maze that has been explored. 
Figure (b) shows the proposed intrinsic reward magnitude for each step in the environment, calculated for both our method and \emph{MAX}. }
\vskip -0.05in
\end{figure}

\setcounter{figure}{6}
\begin{figure*}[ht!]
\centering  
    %\vspace{-3mm}
    \subfigure[Robotic Hand]{
        \label{fig:hand_env}
        \vspace{-1cm}
        \includegraphics[width=.25\linewidth]{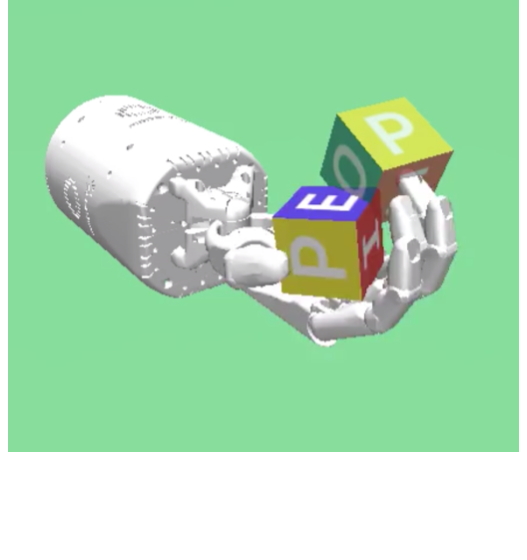}
    }
    \subfigure[Manipulation Task Results]{
        \label{fig:manip_results}
        \includegraphics[width=.71\linewidth]{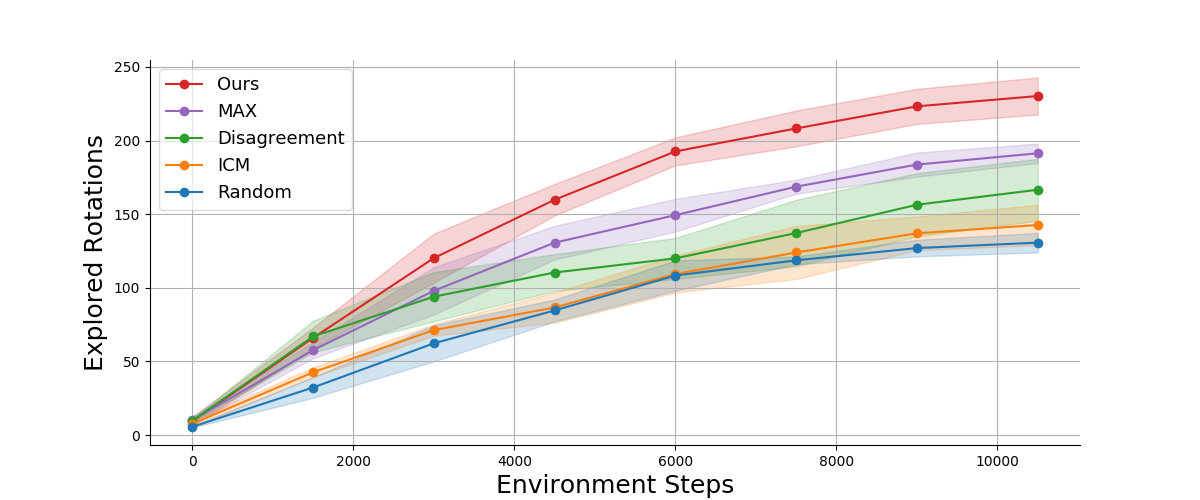}
    }
\vskip -0.1in
\caption{(a) The Robotic Hand task in motion. (b) Performance of each method with mean and $\pm 1$ standard deviation (shaded region) over five seeds. 
$x$-axis is the number of manipulation steps, $y$-axis is the number of rotation states of the block that has been explored.
Our method (red) explores clearly faster than all other methods.}
\vskip -0.05in
\end{figure*}

\subsubsection{Ant Maze Navigation}
Next, we evaluate on the Ant Maze environment. 
In the Ant control task, the agent provides torques to each of the 8 joints of the ant. The provided observation contains the pose of the torso as well as the angles and velocities of each joint. For the purpose of exploration, we place th
e Ant in a U-shaped maze (shown in figure \ref{fig:ant_maze}), where the goal is to reach the end of the maze, discovering all the states. 
The agent's performance is measured by the percentage of the maze explored during evaluation. 
Figure~\ref{fig:nav_results} shows the result of each method over 5 seeds. 
Our agent consistently navigates to the end of the maze faster than the other competing methods. While \emph{MAX}~\citep{shyam2019max} also navigates the maze, the implicit uncertainty modeling scheme in our method allows our agent to better estimate the state novelty, which leads to a considerably faster exploration. 
To see that our agent fully explores the maze, and does not only trace out a single trajectory, we include state visitation diagrams in figures \ref{fig:state_visit1}-\ref{fig:state_visit4}. We see that the agent explores many paths through the maze, and has not left any large portion of the maze unexplored. 

To provide a more intuitive understanding of the effect of an intrinsic reward and how it might correlate to the performance, we also plot in Figure~\ref{fig:ant_int_reward} the intrinsic reward observed by our agent at each exploration step, compared with that observed by the \emph{MAX} agent. For fair comparison we plot the intrinsic reward from eq.(\ref{eq:var_reward}) for both methods. We can see that after step 2K, predictions from the \emph{MAX} ensemble start to become increasingly similar, leading to a decline in intrinsic reward (Fig.~\ref{fig:ant_int_reward}) as well as a slow-down in exploration speed (Fig.~\ref{fig:nav_results}). We hypothesize this is because in a regular ensemble, all members are updating their gradients on the same experiences without an explicit term to match the real posterior, leading to all agents eventually converging to the same representation.
In contrast, our intrinsic reward keeps increasing around step 2K and remains high as we continue to quickly explore new states in the maze, only starting to decline once we have solved the maze at approximately step 5,500.

\subsubsection{Robotic Manipulation}
The final task is an exploration task in a robotic manipulation environment, HandManipulateBlock. As shown in Figure~\ref{fig:hand_env}, a robotic hand is given a palm-sized block for manipulation. The agent has actuation control of the 20 joints that make up the hand, and its exploration performance is measured by the percentage of possible rotations of the cube that the agent performs. This is different from the original goal of this environment since we want to evaluate task-agnostic exploration rather than goal-based policies.
In particular, the state of the cube is represented by Cartesian coordinates along with a quaternion to represent the rotation. We transform the quaternion to Euler angles and discretize the resulting state space by $45$ degree intervals. The agent is evaluated based on how many of the 512 total states are visited. 

This task is far more challenging than previous tasks, having a larger state space and action space. Additionally, states are more difficult to reach than the Ant Maze environment: requiring manipulation of 20 joints instead of 8.  
In order to explore in this environment, an agent must also learn how to rotate the block without dropping it. Figure~\ref{fig:manip_results} shows the performance of each method over 5 seeds. This environment proved very challenging for all methods: none succeeded in exploring more than half of the state space.
Still, our method performs the best by a clear margin.

\subsection{Policy Transfer Experiments}
\label{sec:policy_transfer}
So far, we have demonstrated that the proposed implicit generative modeling of the posterior over dynamic models leads to more effective and efficient pure exploration policies.
While the efficiency of pure exploration is important under sparse reward settings, 
a natural follow-up question is whether a strong pure exploration policy would also be beneficial for downstream tasks where dense rewards are available.
We give an answe
r to this question by performing the following experiments in the widely-used HalfCheetah environment.

\begin{figure*}[ht!]
\centering 
    \subfigure[Policy Transfer Performance with Warm-up]{
        \label{fig:policy_transfer_v1}
        \includegraphics[width=.48\linewidth]{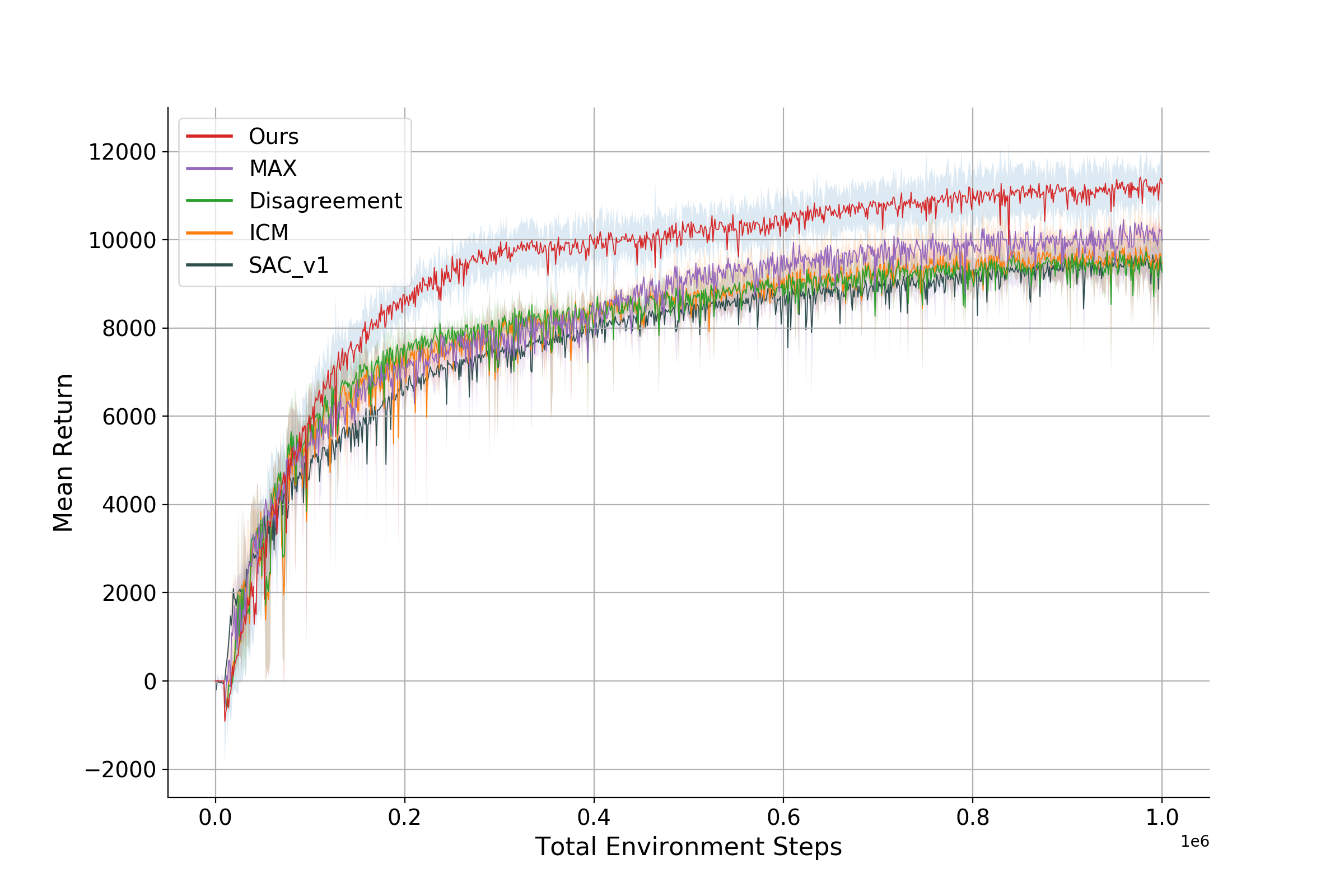}
    }
    \subfigure[Policy Transfer with Varying Exploration Time]{
        \label{fig:policy_transfer_buf}
        \includegraphics[width=.48\linewidth]{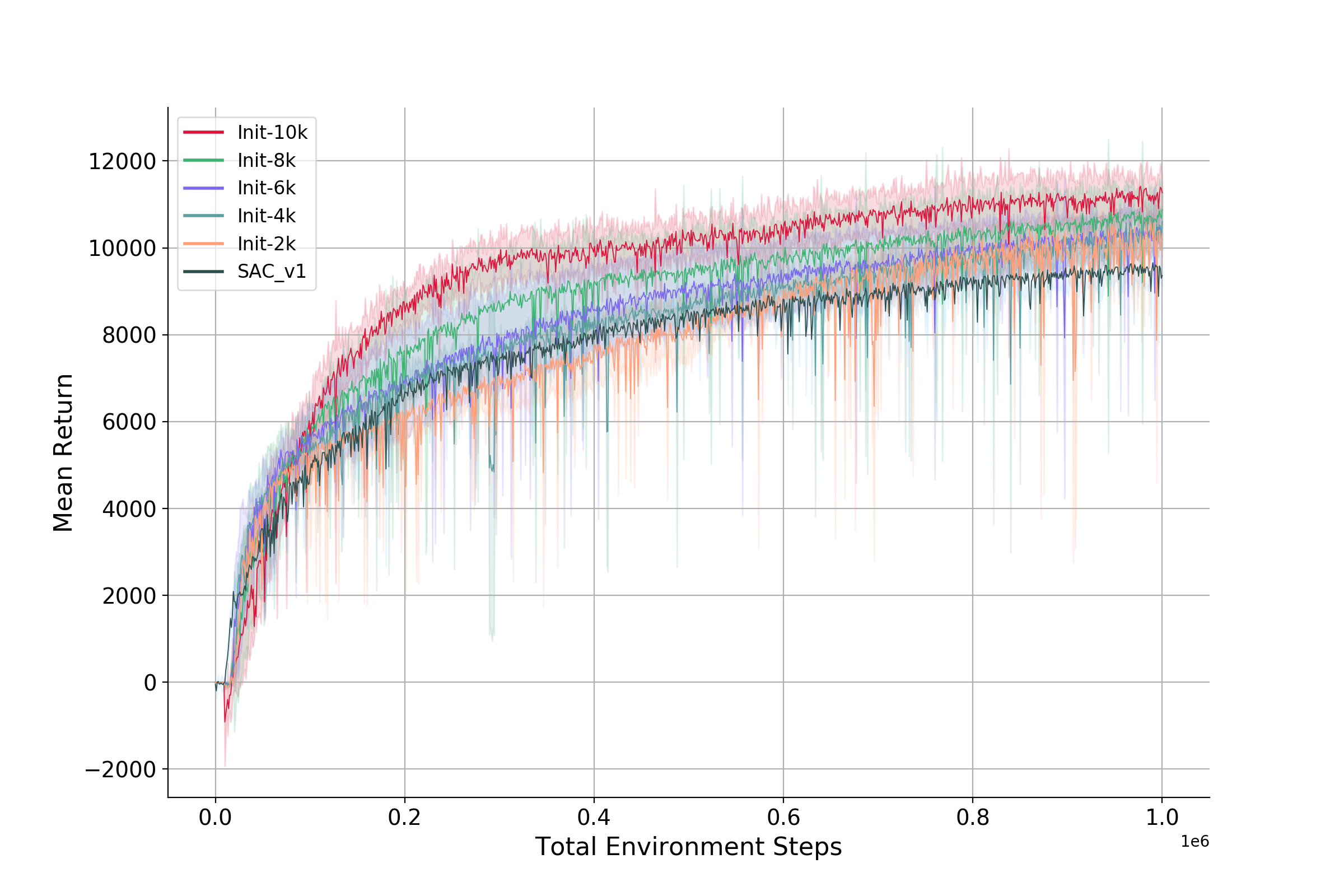}
    }   
    \vskip -0.05in
\caption{Policy transfer results. Figure (a) shows results with policy warm-up on the HalfCheetah environment. We show results for the SAC baseline with uniform warm-up, and SAC with 10K-step pure exploration initialization and warm-up using different intrinsic reward methods: ICM, Disagreement, MAX, and our proposed method (Ours) respectively. To make the comparison fair, agents initialized with pure exploration policies perform 10k less steps during task training than the baseline SAC. Figure (b) 
shows the performance of downstream SAC policy training warmed up with our proposed exploration policy under different numbers of pure exploration steps.
Init-$N$k refers to the SAC agent initialized from our exploration policy which has been trained for $N$-thousand exploration steps.
}
\vskip -0.1in
\end{figure*}

We first train a task-agnostic exploration policy following Alg.~\ref{algo:implicit-active-ex} for 10K environment steps. The trained policy is then used to warm up the (downstream) task policy for an additional 10K environment steps. 
This warm-up stage is followed by standard training of the task policy using external rewards. 
Warm-up periods are often used as an initial exploration stage, to collect enough data for meaningful off-policy updates. 
The baseline SAC is SAC v1 \citep{haarnoja2018soft}, where the warm-up stage consists of taking uniformly random actions for the first 10K steps, before performing any parameter updates. 
Given that our trained exploration policies explore much more efficiently than acting randomly, we can examine if warming up SAC with our exploration policy offers a benefit over the standard uniform warm-up strategy.%, when applied for downstream task learning. 

In particular, we first train a pure exploration policy for 10K steps, for each method on HalfCheetah. 
We then freeze the parameters of the pure exploration policy, and use them to initialize a new agent in a HalfCheetah environment where the external reward is known. 
We then warm up the agent, taking actions and collecting data according to the newly initialized policy. 
After 10K steps of warm-up we begin training as normal, with respect to the external reward. 
We 
evaluate this procedure that we call ``policy transfer", by comparing the performance of SAC at 1M steps, after the task policy has been warmed up using exploration policies trained by \emph{MAX}, \emph{ICM}, \emph{Disagreement}, and \emph{Ours} respectively. 
We also include SAC v1 (with a uniform warm-up strategy) as a baseline. 
For the training of the task policy, we follow the recommended settings for HalfCheetah given in the original SAC v1 method. 
In the supplementary material, we detail our hyper-parameter choices, as well as show that our choice of hyper-parameters does not unfairly favor our method. 

Figure \ref{fig:policy_transfer_v1} shows the performance of all compared methods on HalfCheetah. We can see that the comparatively small number of initial steps spent on pure exploration pays off when the agent switches 
to the downstream task.
Even though SAC is widely regarded as a strong baseline with a maximum entropy-based exploration mechanism, all intrinsic reward methods are able to improve the baseline more or less, by introducing a pure exploration stage before standard training of SAC. We also observe that the stronger the pure exploration policy is, the more it can improve the training efficiency of the downstream task. Task policies initialized with our exploration policy (\emph{Ours}) still perform the best with a clear margin.  

We also conduct an ablation study to better understand the relationship between the number of steps used to train the exploration policy, and the improvement it brings to downstream task training. 
We compare multiple variants of \emph{Ours} in Figure~\ref{fig:policy_transfer_v1}, with different numbers of pure exploration steps: 2K, 4K, 6K, 8K, and 10K steps.
As shown in Figure~\ref{fig:policy_transfer_buf}, with only 2K or more steps of initial pure exploration, our approach improves upon the SAC baseline in the downstream task. 
The longer our exploration policy is trained, the more beneficial it is to the training of the downstream task. We note that by using an exploration policy trained for just 4K steps, our agent performs strictly better on the downstream task than the SAC baseline.  

Our next study shows that even in a setting without a warm-up stage, initializing the task policy with a pure exploration policy still benefit the downstream task learning. In figure \ref{fig:sac_no_warmup}
we show an evaluation of SAC on the HalfCheetah environment without the warm-up stage. We report performance of the SAC baseline, as well as SAC initialized with exploration policies trained by \emph{MAX}, \emph{ICM}, \emph{Disagreement}, and \emph{Ours}. We can see that without initial r
andom exploration, the performance of SAC suffers dramatically. Policies initialized with pure exploration policies outperform the baseline following the same trend as in the setting with warm-up. \emph{Ours} still performs the best with a clear margin.
% In figure \ref{fig:sac_no_warmup} we show an evaluation of SAC initialized with a pure exploration policy, and trained for 1M steps with \emph{MAX}, \emph{ICM}, \emph{Disagreement}, and \emph{Ours} respectively, without a warm-up stage. As a baseline, we also show SAC trained without a warm-up stage. We can see that SAC depends heavily on the warm-up stage for downstream performance; without initial random exploration, its performance suffers dramatically. Policies initialized with pure exploration however, are less sensitive to the lack of warm-up stage, and are all able to improve performance over the baseline. 

\renewcommand{\thefigure}{9}
\begin{figure}[htb]
\vskip -0.2in
\centering 
    \includegraphics[width=1.0\linewidth]{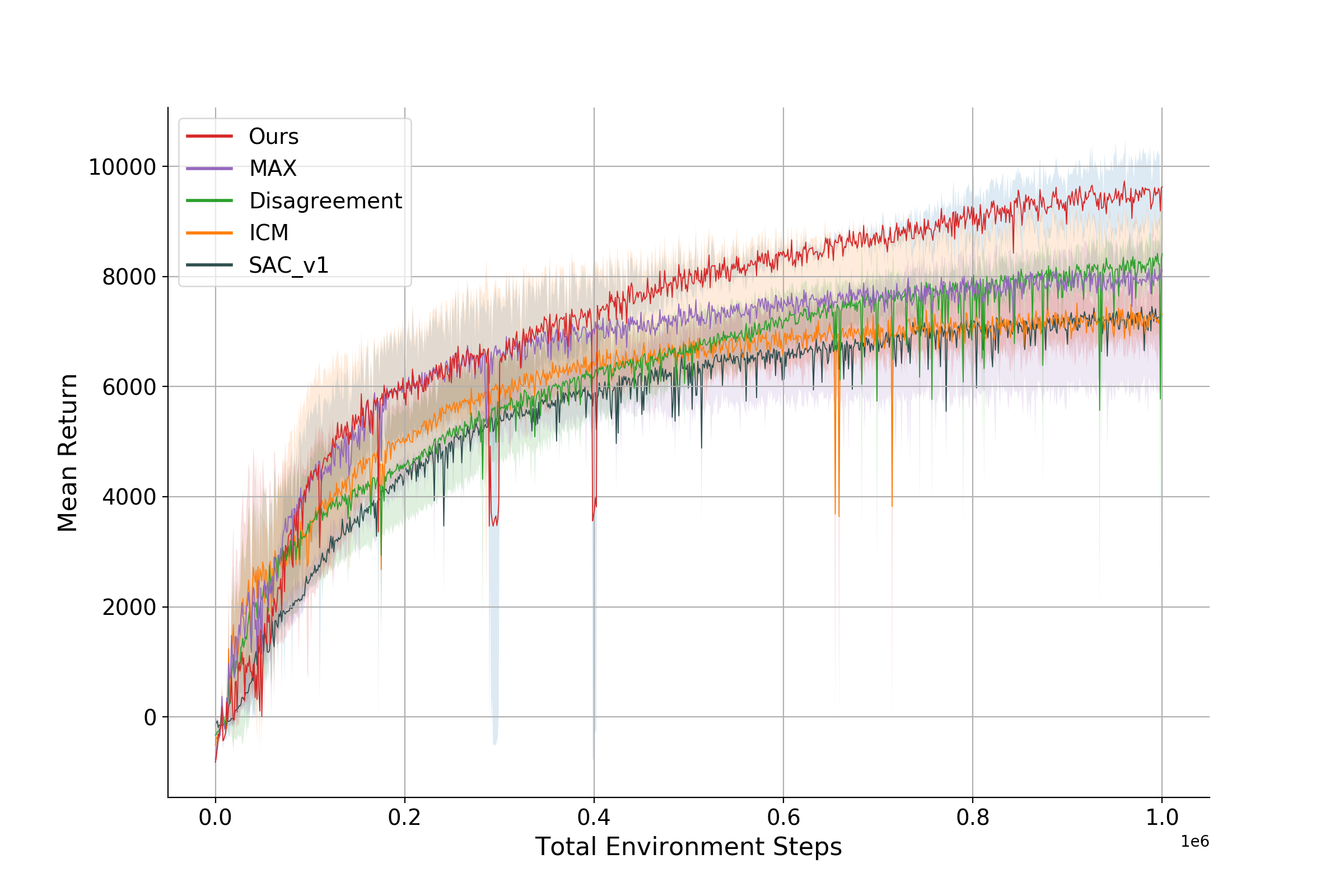}
\vskip -0.15in
\caption{Policy transfer results without warm-up stage, on the HalfCheetah environment. We show our results for the SAC baseline without uniform warm-up, as well as an SAC initialized with exploration policies trained by \emph{MAX}, \emph{ICM}, \emph{Disagreement}, and \emph{Ours}, respectively}
    \label{fig:sac_no_warmup}
\vskip -0.2in
\end{figure}

%Improving performance without using a warm-up stage is indicative of the ability of the pure-exploration policy to collect diverse data. 
%The role of the warm-up stage is to initialize the agent with sufficiently diverse data such that off-policy updates do not immediately overfit to the environment local to the agent. 
%The decrease in the performance of SAC without a warm-up stage is likely from its inability to collect meaningful data at the outset. In contrast, 
%By initializing with a pure exploration policy, the agent already knows how to collect diverse data, and therefore is less prone to overfit to the initially collected data. The better the initial exploration policy is, the more efficiently the agent is able to collect diverse data for task learning. 
% In safety-critical applications where collecting data through random actions is unwise, initializing with a pure exploration policy is a viable alternative. 

\section{Conclusion and Future Work}
In this work, we introduced a new method for representing the agent's uncertainty of the environment dynamics. Utilizing amortized SVGD, we learned an approximate posterior over dynamics models. We use this approximate posterior to formulate an intrinsic reward based on the uncertainty estimated from samples of this distribution, enabling efficient exploration in difficult environments, 
Future work includes investigating 
the efficacy of learning an approximate posterior of the agent's value or policy model, as well as more efficient sampling techniques to reduce the computational cost inherent to many model-based algorithms. We would also investigate principled methods of combining intrinsic and external rewards, and how different exploration policies influence the downstream task performance. 

\subsection*{Acknowledgements}
Neale Ratzlaff and Li Fuxin were partially supported by the Defense Advanced Research Projects Agency (DARPA) under Contract No. N66001-17-12-4030, HR001120C0011 and HR001120C0022. Any opinions, findings and conclusions or recommendations expressed in this material are those of the author(s) and do not necessarily reflect the views of DARPA.
\bibliographystyle{icml2020}
\bibliography{icml2020}

\onecolumn

%\icmltitle{Supplementary Material to Implicit Generative Modeling for Efficient Exploration}
%\icmltitlerunning{Implicit Generative Modeling for Efficient Exploration}

\appendix
\section{Supplementary Material}

\subsection{Exploration Environment Implementation Details}
Here we describe in more detail the various implementation choices we used for our method as well as for the baselines. 

\noindent {\bf Toy Chain Environment}

The chain environment is implemented based on the NChain-v0 gym environment. We alter NChain-v0 to contain 40 states instead of 10 to reduce the possibility of solving the environment with random actions. We also modify the stochastic 'slipping' state behavior by fixing the behavior of the states respect to reversing an action. For both our method and \emph{MAX}, we use ensembles of 5 deterministic neural networks with 4 layers, each is 256 units wide with tanh nonlinearities. As usual, our ensembles are sampled from the generator at each timestep, while \emph{MAX} uses a static ensemble. We generate each layer in the target network with generators composed of two hidden layers, 64 units each with ReLU nonlinearities. Both models are trained by minimizing the regression loss on the observed data. We optimize using Adam with a learning rate of $10^{-4}$, and weight decay of $10^{-6}$. We use Monte Carlo Tree Search (MCTS) to find exploration policies for use in the environment. We build the tree with 25 iterations of 10 random trajectories, and UCB-1 as the selection criteria. Crucially, when building the tree, we query the dynamic models instead of the simulator, and we compute the corresponding intrinsic reward. For intrinsic rewards, \emph{MAX} uses the Jensen Shannon divergence while our method uses the variance in the predictions within the ensemble. After building the tree we take an action in the real environment according to our selection criteria. There is a small discrepancy between the numbers reported in the MAX paper for the chain environment. This is due to using UCB-1 as the selection criteria instead of Thompson sampling as used in the \emph{MAX}. 
We take actions in the environment based on the children with the highest value. The tree is then discarded after one step, after which, the dynamic models are fit for 10 additional epochs. 

\noindent {\bf Continuous Control Environments}

For each method where applicable, we use the method-specific hyperparameters given by the authors. Due to experimenting on potentially different environments, we search for a suitable learning rate which works the best for each method across all tasks. The common details of each exploration method are as follows. Each method uses (or samples) an ensemble of dynamic models to approximate environment dynamics. An ensemble consists of 32 networks with 4 hidden layers, 512 units wide with ReLU nonlinearities, except for \emph{MAX} which uses swish\footnote{Swish refers to the nonlinearity proposed by~\citep{ramachandran2017searching} which is expressed as a scaled sigmoid function: $y = x + sigmoid(\beta x)$}. \emph{ICM}, \emph{Disagreement}, and our method use ensembles of deterministic models, while \emph{MAX} uses probabilistic networks which output a Gaussian distribution over next states. The approximate dynamic models (ensembles/generators) are optimized with Adam, using a minibatch size of 256, a learning rate of $1.0^{-4}$, and weight decay of $1.0^{-5}$.   

For our dynamic model, each layer generator is composed of two hidden layers, 64 units wide and ReLU nonlinearity. The output dimensionality of each generator is equal to the product of the input and output dimensionality of the corresponding layer in the dynamic model. To sample one dynamic model, each generator takes as input an independent draw from $z \sim \mathcal{Z}$ where $\mathcal{Z} = \mathcal{N}(\mathbf{0}^{32}, \mathbf{1}^{32})$. We sample ensembles of a given size $m$ by instead providing a batch $\{z\}_{i=1}^m$ as input. To train the generator such that we can sample accurate transition models, we update according to equation (4) in the main text; we compute the regression error on the data, as well as the repulsive term using an appropriate kernel. For all experiments we use a standard Gaussian kernel 
%$k(\theta, \theta_i) = \exp\left(-\frac{1}{h}||\theta - \theta_i||^2_2\right)$ 
$K(f_{\theta_i},f_{\theta_j}) = \exp{(-d(f_{\theta_i}, f_{\theta_j})/h)}$, where
$d(f_{\theta_i}, f_{\theta_j}) = \frac{1}{n}\sum\limits^n_{l=1}\|f_{\theta_i}(x_l)-f_{\theta_j}(x_l)\|^2_2$ for a training batch $\{x_l\}_{l=1}^n$. Where $h$ is the median of the pairwise distances between sampled particles $\{f_\theta\}_{i=1}^m$. Because we sample functions $f_{\vec{\theta}}$ instead of data points, the pairwise distance is computed by using the likelihood of the data $\vec{x}$ under the model: $\log f_{\vec{\theta}}(\vec{x})$.  

For \emph{MAX}, we use the code provided from~\citep{shyam2019max}\footnote{https://github.com/nnaisense/max}. Each member in the ensemble of dynamic models is a probabilistic neural network that predicts a Gaussian distribution (with diagonal covariance) over the next state. The exploration policy is trained with SAC, given an experience buffer of rollouts $\Bar{D} = \{s, a, s'\} \cup R\pi$ performed by the dynamic models, where $R_\pi$ is the intrinsic reward: the Jensen-Renyi divergence between next state predictions of the dynamic models. The policy trained with SAC acts in the environment to maximize the intrinsic reward, and in doing so collects additional transitions that serve as training data for the dynamic models for the subsequent training phase. 

For \emph{Disagreement}~\citep{pathak2019disagreement}, we implement this method under the MAX codebase, following the implementation given by the authors\footnote{https://github.com/pathak22/exploration-by-disagreement}. The intrinsic reward is formulated as the predictive variance of the dynamic models, where the models are represented by a bootstrap ensemble. In this work, we report results using two versions of this method. The proposed intrinsic reward specifically is formulated in a manner quite similar to our own, however, a fixed ensemble is used instead of a distribution for the approximate posterior. 
In section~\S4 of the main text we report results of \emph{Disagreement} only using its intrinsic reward, instead of the full method, which makes use of a differentiable reward function and treats the reward as a supervised learning signal. We examine these methods separately because we are testing the effects of intrinsic rewards, as well as the form of the approximate dynamic model e.g. sampling vs fixed ensembles. The differentiable reward function is orthogonal to this effort. 
Nonetheless, in the next section~\S\ref{sec:disagreement_results} we report results using the full method of \emph{Disagreement}, on each continuous control experiment. 

\subsection{Extended Disagreement Results}
\label{sec:disagreement_results}
\begin{figure}[ht!]
\centering 
    \subfigure[Acrobot]{
        \label{fig:app_acrobot}
        \includegraphics[width=.31\linewidth]{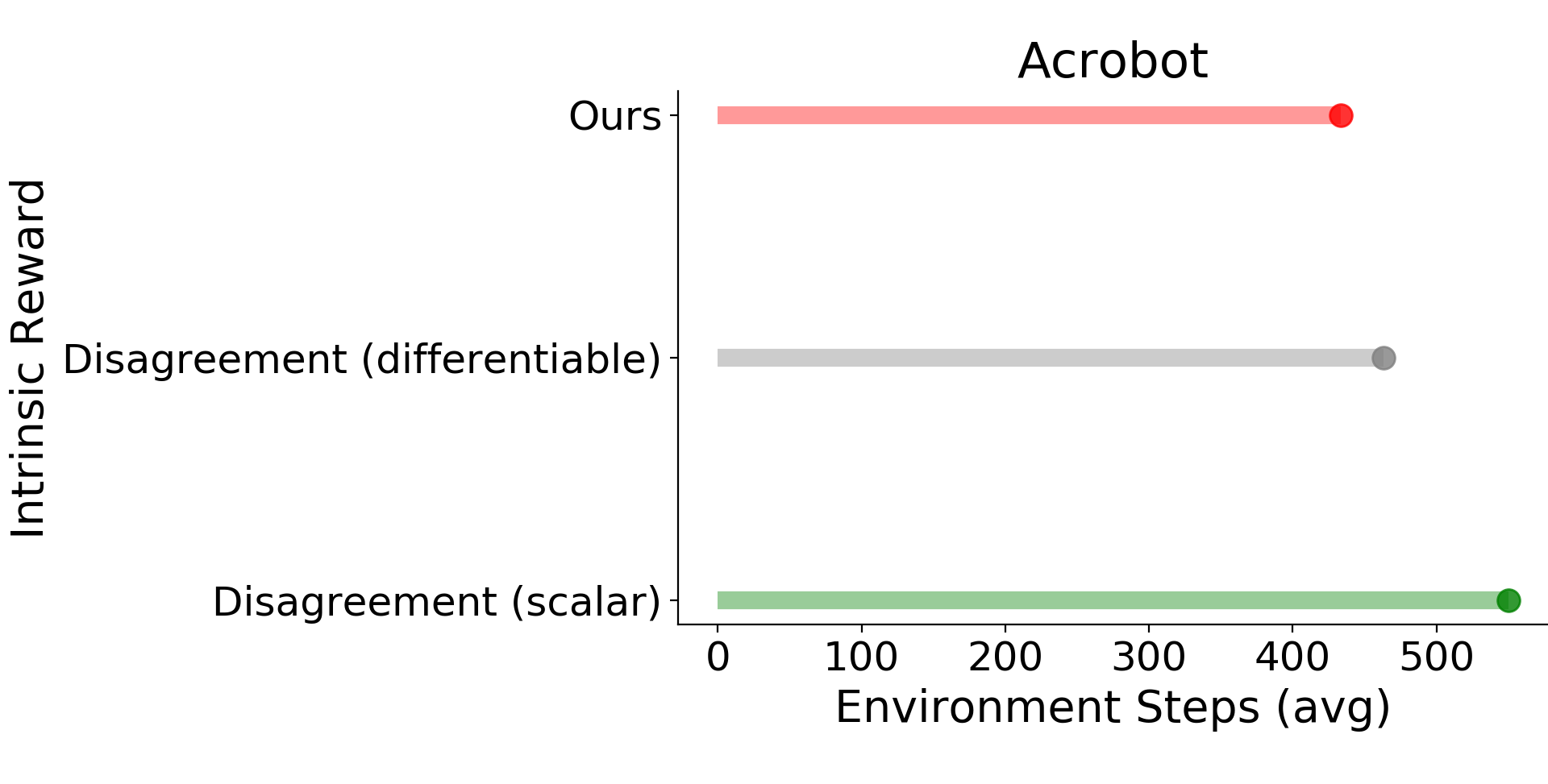}
    }
    \subfigure[Ant Maze]{
        \label{fig:app_ant}
        \includegraphics[width=.31\linewidth]{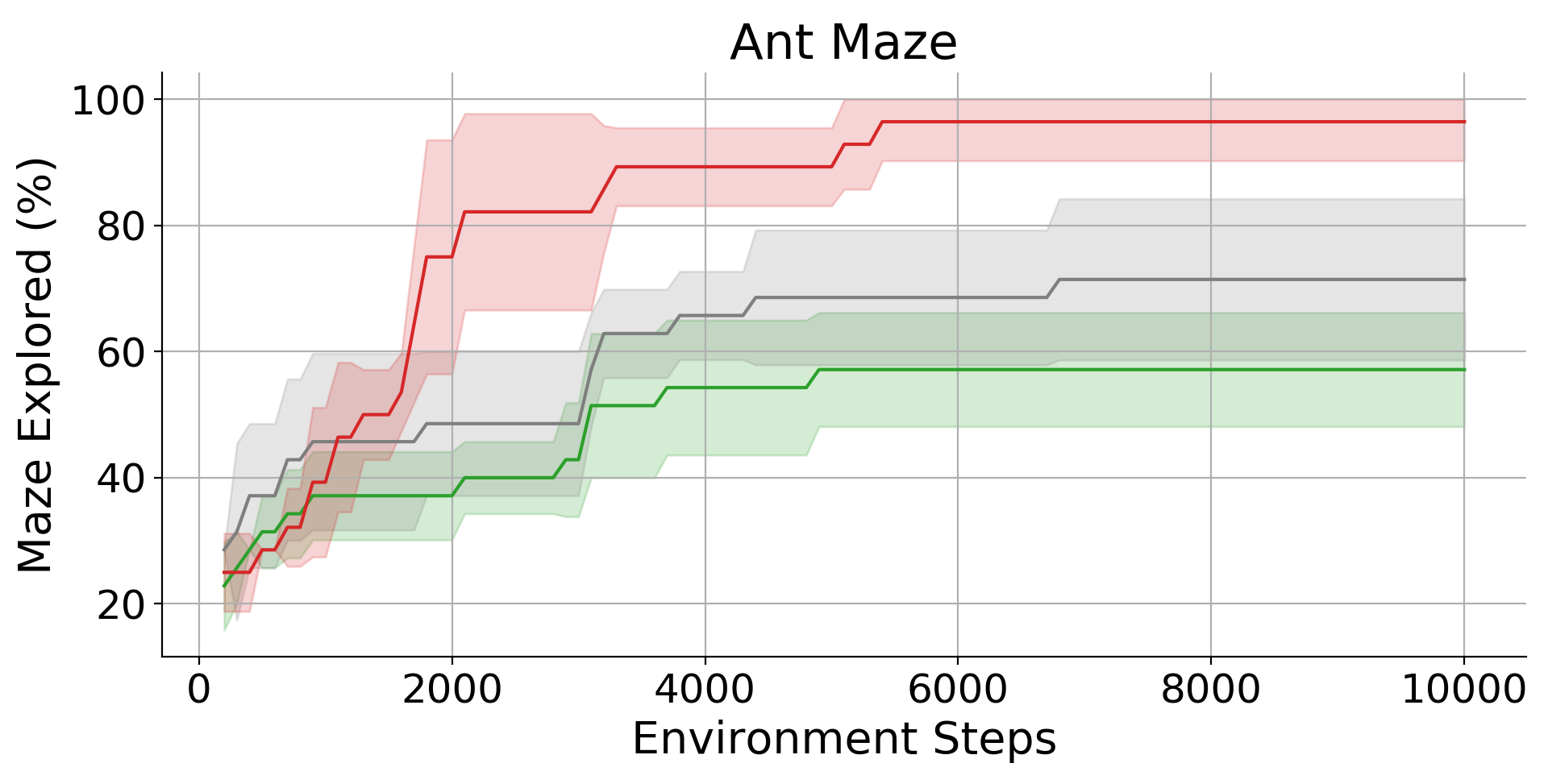}
    }
    \subfigure[HandManipulateBlock]{
        \label{fig:app_block}
        \includegraphics[width=.31\linewidth]{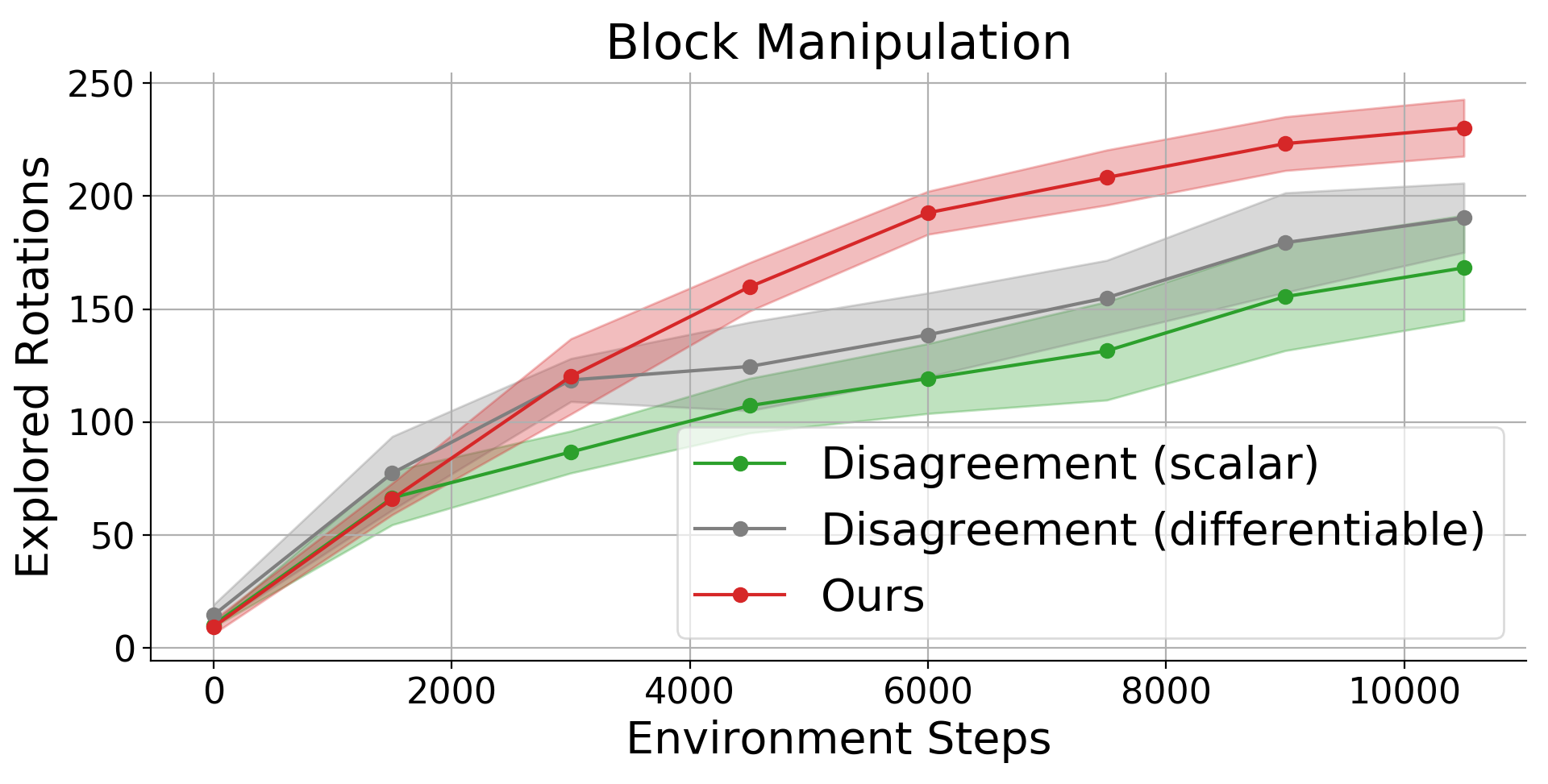}
    }
    \caption{Results for the full \emph{Disagreement} method including the differentiable reward function on the Acrobot (a), Ant Maze (b), and HandManipulateBlock (c) environments.}
\end{figure}

Here we report additional comparisons with \emph{Disagreement} -- including the original policy optimization method with a differentiable reward function \citep{pathak2019disagreement}.
We repeat our pure exploration experiments, comparing our method to both disagreement purely as an intrinsic reward, as well as the full method using the differentiable reward function for policy optimization. 
Figures~\ref{fig:app_acrobot},~\ref{fig:app_ant}, and~\ref{fig:app_block} show results on the Acrobot, Ant Maze, and Block Manipulation environments, respectively. In each figure, lines correspond to the mean of three seeds, and shaded regions denote $\pm$ one standard deviation. In each experiment, we can see that treating the intrinsic reward as a supervised loss (gray) improves on the baseline scalar-valued disagreement intrinsic reward (green). However, our method (red) remains the most sample efficient in these experiments.

\subsection{Comparison to other exploration methods}
Here we compare our method to two other representative exploration methods. Parameter Space Noise for Exploration (PSNE) \cite{plappert2017parameter} adds parametric noise to the weights of an agent, similar to \cite{fortunato2017noisy}. 
The noise parameters are learned by gradient descent, and the additional stochasticity in the induced policy is responsible for increased exploration ability. 
Random Network Distillation (RND) is another well-known method \cite{burda2018exploration} that introduces a randomly initialized function $f: S \rightarrow \mathbb{R}^k$ which maps states $s$ to a k-dimensional vector, similar to \cite{osband2018prior}. 
A second function $\hat{f}: S \rightarrow \mathbb{R}^k$ is trained to match the predictions given by $f$. 
The prediction error $\hat{f(s)} - f(s)$ is used as an exploration bonus to the reward during training, similar to the psuedo-count based exploration bonus in \cite{bellemare2016unifying}. 
RND has been shown to be a strong baseline for both task-specific environments and pure exploration.

\begin{figure}[ht!]
\centering 
    \subfigure[Baseline comparison on HalfCheetah]{
        \label{fig:comp_cheetah}
        \includegraphics[width=.48\linewidth]{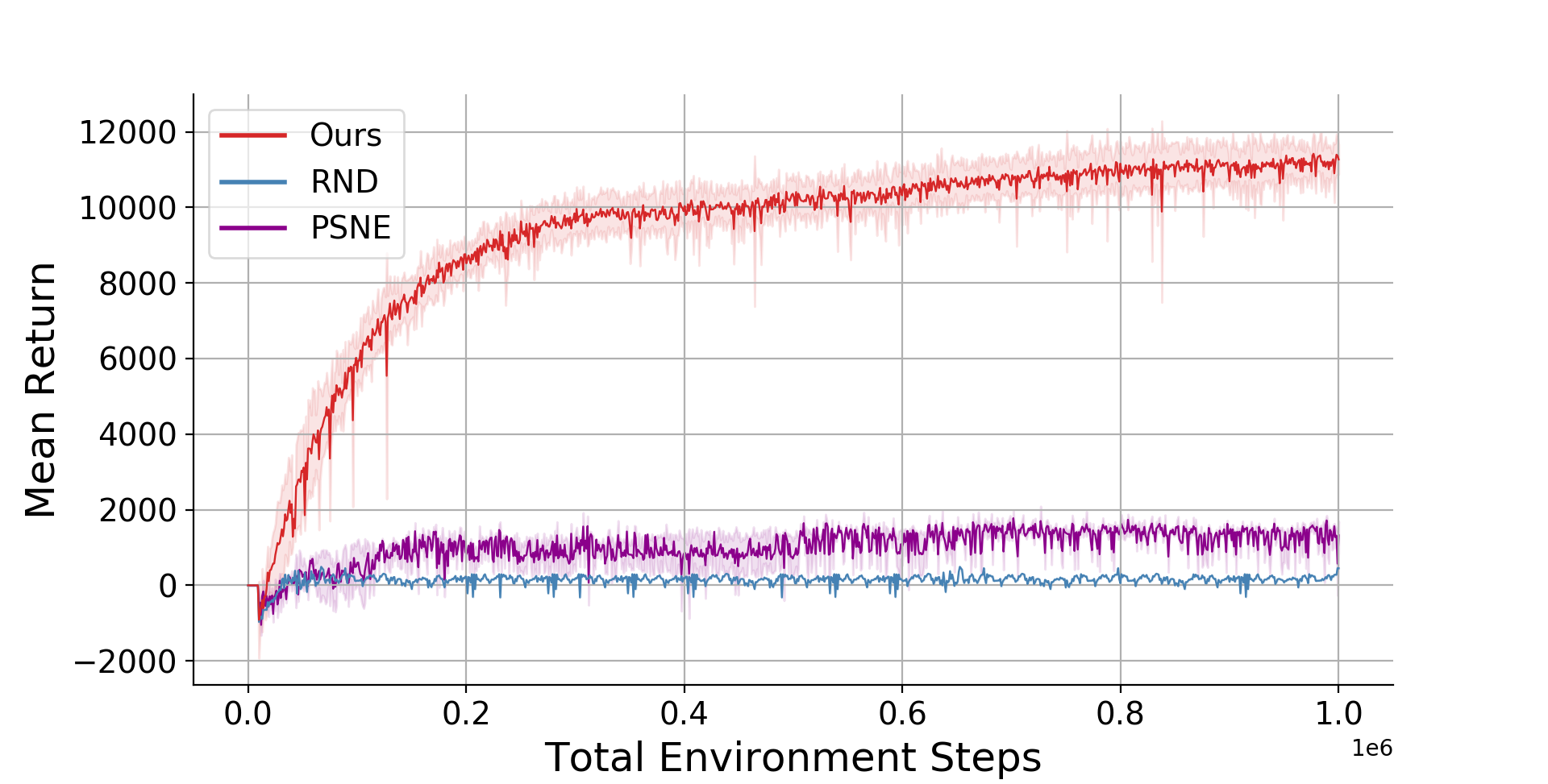}
    }
    \subfigure[Baseline comparison on Ant Maze]{
        \label{fig:comp_ant}
        \includegraphics[width=.48\linewidth]{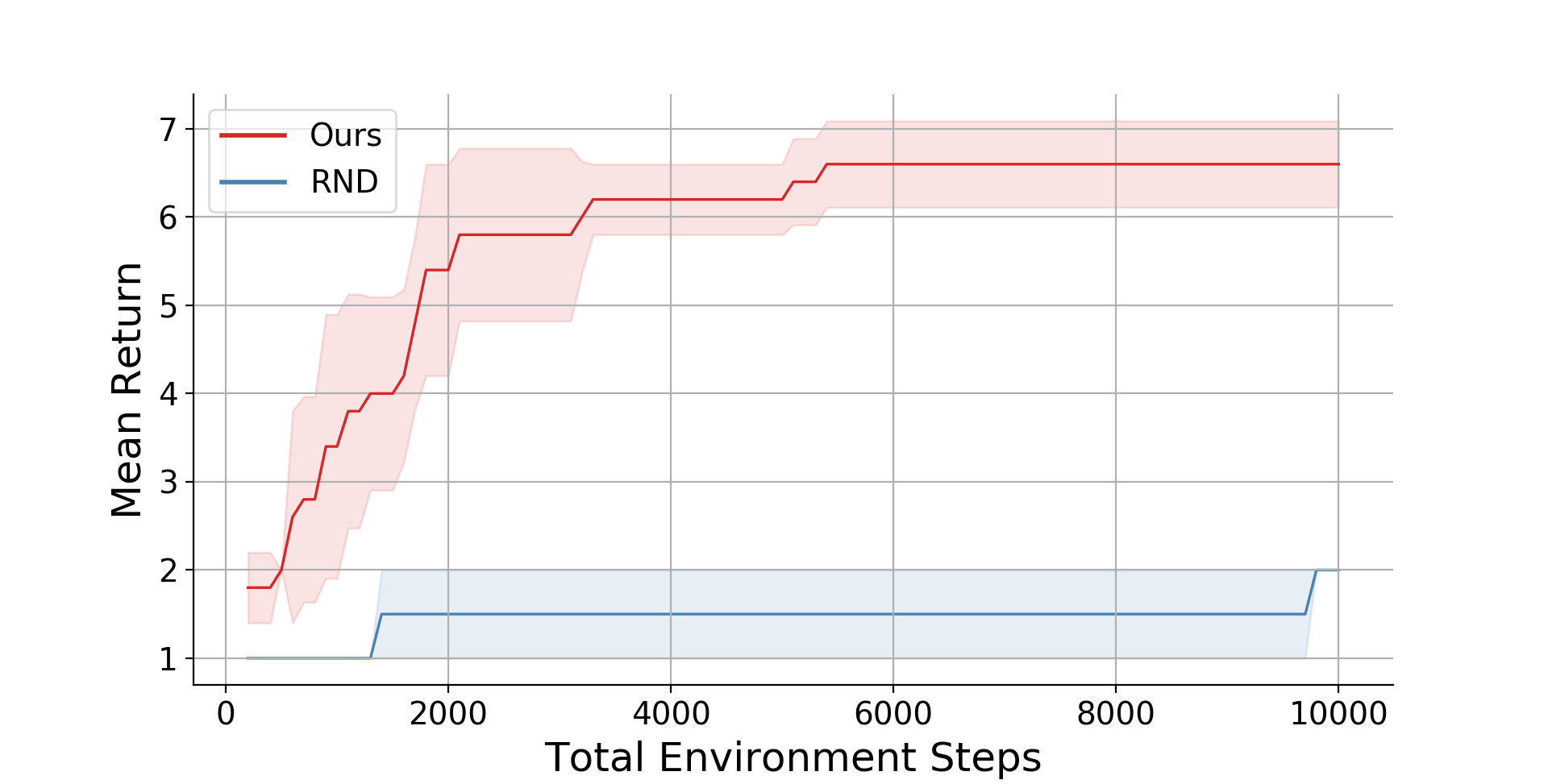}
    }

    \caption{Comparison with RND and PSNE on HalfCheetah with external reward (a), and a pure exploration comparison with RND on Ant Maze (b).}
\end{figure}

In Figure~\ref{fig:comp_cheetah}, we first compare our method with PSNE and RND on the HalfCheetah environment, as both can used to learn task-specific policies. For both methods, we use the author provided codes to run our experiments. Because RND is initially designed for discrete actions, we modify the policy to handle continuous action spaces. However, we were unable to recover the reported results from PSNE using the provided code\footnote{For PSNE we used the code at https://github.com/openai/baselines.}. In Figure~\ref{fig:comp_ant}, we further compare with RND in the pure exploration setting. We omit PSNE from this experiment, as PSNE does not have intrinsic reward, or another mechanism that can be directly used for pure exploration in the Ant Maze environment. For Ant Maze, each method runs for 10k steps for pure exploration, without external reward.

These methods lack an explicit model of the environment dynamics. It has been shown many times that model-based methods have a considerable advantage in sample efficiency. RND in particular, takes hundreds of millions of environment steps to achieve its performance. We show in Figure \ref{fig:comp_cheetah} and \ref{fig:comp_ant} that our method enables superior downstream task performance, and better sample efficiency in exploration, respectively.  

\subsection{Policy Transfer Implementation Details}
Here we describe in detail the specific settings and design choices used for the policy transfer methods and environments. 

\par\noindent\textbf{Policy Transfer with Warm-up}\\
The exploration policies for our method, \emph{MAX}, \emph{ICM}, and \emph{Disagreement} were trained exactly as in the pure exploration experiments. We trained each exploration policy for 10k steps. We then initialize a new SAC agent with the exploration policy. This agent is initialized within the HalfCheetah environment that includes the external reward. Given that the new agent has an empty replay buffer, we perform a warm-up stage to collect initial data before performing any parameter updates. We collect this initial data by rolling out the pure exploration policy for 10K steps, and storing the observed transitions in the fresh agent's replay buffer. Note that the policy is frozen during the warm-up. After this initial warm-up stage, we allow the fresh agent to train as normal for 1M steps (including the steps taken during warm-up and pure exploration), with respect to the external reward. 

\par\noindent\textbf{Policy Transfer Without Warm-up}\\
The policy transfer experiments without the warm-up stage are similar in that the pure exploration polices are trained for 10K steps, agnostic of the downstream task, then frozen. However, instead of training a new SAC agent on transitions obtained via a warm-up stage, we only transfer the parameters of the pure exploration policy to the fresh SAC agent. 
Then the transition buffer is cleared, and the agent is trained in the standard setting with external reward for 1M steps (including the steps already taken during pure exploration). 

\par\noindent\textbf{Hyper-parameter Comparison}\\
We show the hyper-parameters that we use for each pure exploration method, as well as the SAC baseline in table \ref{tab:pt_hp}. 

\begin{table}[]
\centering
\begin{tabular}{|c|c|c|c|l|c|}
\hline
              & Ours & MAX  & Disagreement & ICM  & SAC Baseline  \\ \hline
Learning Rate & 1e-3 & 1e-3 & 1e-3         & 1e-3 & 3e-4 \\ \hline
Batch Size    & 4096 & 4096 & 4096         & 4096 & 256  \\ \hline
Alpha         & 0.02 & 0.02 & 0.02         & 0.02 & 1    \\ \hline
Hidden Size   & 256  & 256  & 256          & 256  & 256  \\ \hline
Gamma         & 0.99 & 0.99 & 0.99         & 0.99 & 0.99 \\ \hline
Tau           & 5e-3 & 5e-3 & 5e-3         & 5e-3 & 5e-3 \\ \hline
Reward Scale  & 1    & 1    & 1            & 1    & 5    \\ \hline
\end{tabular}
\caption{List of SAC Hyper-parameters used with each method for pure exploration and policy transfer experiments}
\label{tab:pt_hp}
\end{table}

For the SAC baseline we use the hyper-parameters given by the authors \citep{haarnoja2018soft} for HalfCheetah. When training the pure exploration policies, we use the hyperparameters given by~\citet{shyam2019max}. To ensure that we are using the best set of hyper-parameters for each method, we have run baseline SAC with the hyper-parameters used in our method, but they did not perform better than the ones given by the authors. We show in table \ref{tab:hp_comparison} that baseline SAC with and without a warm-up stage, performs best when using the hyper-parameters given by the authors, rather than those we selected for training pure exploration policies. In a similar vein, we can see that our method benefits from using the hyper-parameters given by~\citet{shyam2019max}. Note that when training on the task policy, we always use the original SAC v1 hyper-parameters. 

\begin{table}[]
\centering
\begin{tabular}{|l|c|c|c|c|}
\hline
                         & Ours         & Ours (warm-up)& SAC Baseline & SAC Baseline (warm-up) \\ \hline
Hyperparameters\textsubscript{SAC v1} & 9701 $\pm$ 210 & 10999 $\pm$ 355 & 7273 $\pm$ 537 & 9363 $\pm$ 277 \\ \hline
Hyperparameters\textsubscript{exp}    & 9631 $\pm$ 559 & 11269 $\pm$ 395 & 6902 $\pm$ 696 & 9321 $\pm$ 677 \\ \hline
\end{tabular}
\caption{Comparison of hyper-parameter choices for both our method and baseline SAC, with and without a warm-up stage. We show the final performance of each method after 1M steps, over 3 trials. Hyper-parameters\textsubscript{SAC v1} refers to the hyper-parameters given in \citep{haarnoja2018soft} for HalfCheetah, and Hyperparameters\textsubscript{exp} refers to the hyper-parameters given in \citep{shyam2019max}.}
\label{tab:hp_comparison}
\end{table}

\end{document}